\documentclass{article}

\usepackage{natbib}
\usepackage{edisonstyle}
\providecommand{\doi}[1]{\href{https://doi.org/#1}{doi:\nolinkurl{#1}}}

\let\cite\citep

\edisontitle{\textit{BixBench3}: Benchmarking AI agents on research-study-scale computational biology tasks}
\edisonauthors{Zane Koch\textsuperscript{1}, Asmamaw T. Wassie\textsuperscript{1}, Javier Valdes-Aleman\textsuperscript{1}, Jason Lee\textsuperscript{1}, Michaela M. Hinks\textsuperscript{1}, Samuel G. Rodriques\textsuperscript{1}, Andrew D. White\textsuperscript{1}, Jon M. Laurent\textsuperscript{1,*}}
\edisonaffil{\textsuperscript{1}Edison Scientific, Inc., San Francisco, CA, USA \\ \textsuperscript{*}Correspondence to Jon M. Laurent at \texttt{jon@edisonscientific.com} \\ \textit{Preprint. August 27, 2026.}}
\edisonabstract{Artificial intelligence (AI) promises to accelerate biological research by automating computational analyses. Yet the ability of AI agents to carry out computational biology at the scale of complete research studies has not been systematically evaluated. Here we introduce \textit{BixBench3}, a benchmark that measures the capacity of AI agents to process raw biological data through to scientific results. We designed \textit{BixBench3} tasks to mirror the delegation of work from a scientist to an agent: the scientist chooses the research question and high-level methods, then delegates implementation of all analyses to the agent. In each task, an agent receives a research objective, methodological guidance, and raw data derived from a published scientific study, and must execute a sequence of analyses to achieve the research objective. The data artifacts resulting from these analyses – such as peak call matrices or differential expression tables – are programmatically graded against the corresponding artifacts generated and reported in the original study. Across 20 \textit{BixBench3} tasks encompassing the generation of 138 unique artifacts, we find that 13 frontier large language models (LLMs) achieve scores ranging from 0.00 for Gemini 3.1 Flash Lite to 0.48 for GPT~5.6~Sol. Agents perform worse on tasks with larger raw datasets (0.36 on tasks with $<100$~GB versus 0.10 on tasks with $>100$~GB) and on analyses requiring more sequential steps (0.36 at 1--2 steps vs 0.24 at 3+). On average, agents use 6.8 hours, 102 million tokens, and \$43 to complete each task, with the longest attempts consuming 24 hours, 1.07 billion tokens, and \$525. Notably, the highest-scoring agents used fewer tokens and were cheaper than less performant options. These results reveal that LLMs vary substantially in their ability to (1) execute multiple sequential analysis steps coherently, (2) manage large quantities of raw data, and (3) work across scientific domains.}
\edisonlogo{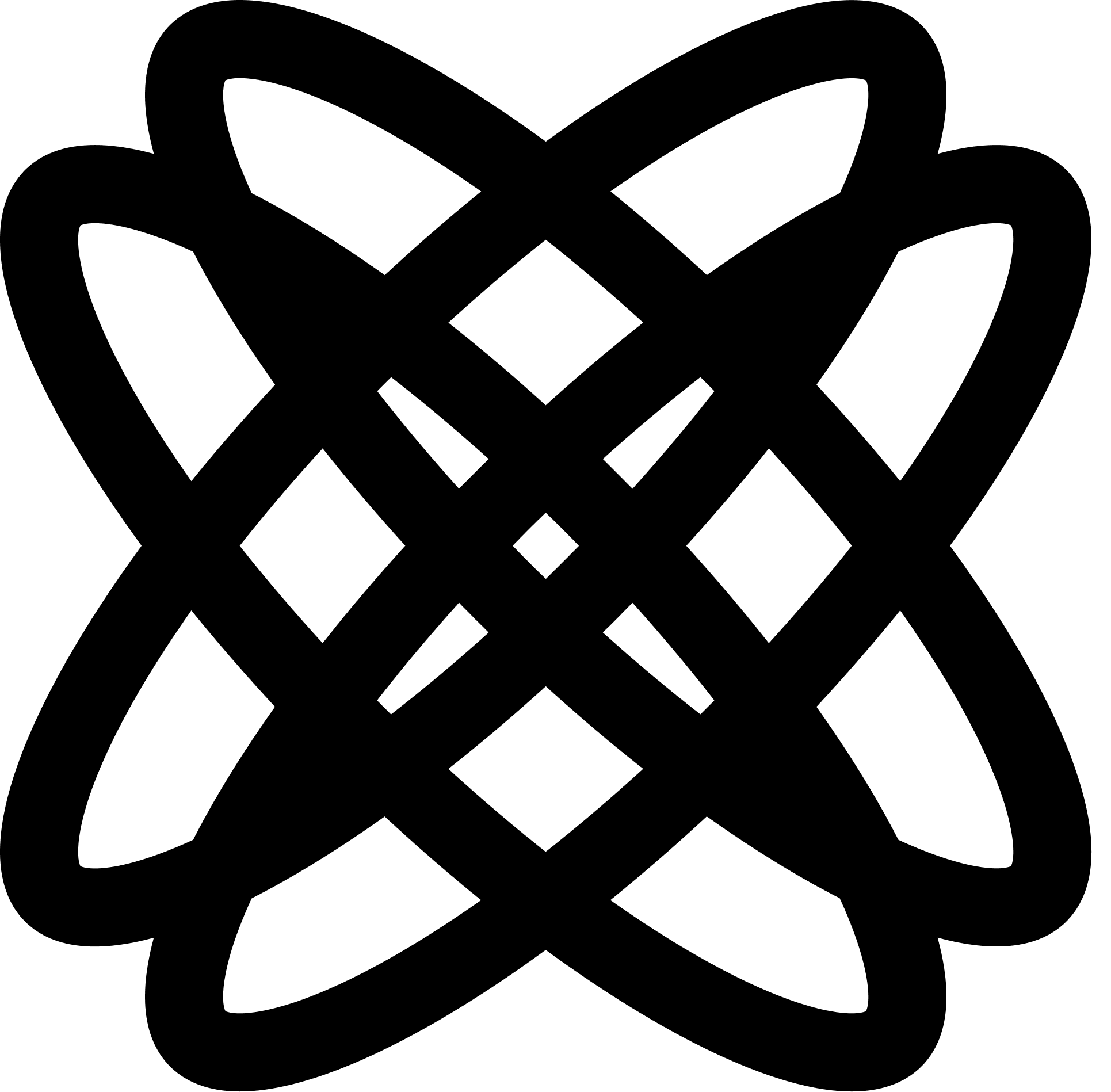}

\usepackage{microtype}
\usepackage{listings}
\graphicspath{{figures/}}


\hypersetup{
	pdftitle={BixBench3: Benchmarking AI agents on research-study-scale computational biology tasks},
	pdfauthor={Zane Koch, Asmamaw T. Wassie, Javier Valdes-Aleman, Jason Lee, Michaela M. Hinks, Samuel G. Rodriques, Andrew D. White, Jon M. Laurent}
}

\begin{document}
\pagestyle{plain}

\makeedisonheader

\section{Introduction}
\label{sec:intro}

LLM-based agents are increasingly being deployed to perform scientific research. Recent agents combine large language models with code execution, retrieval, and domain-specific tools to carry out parts of research workflows, rather than isolated question answering. In biology, Robin drove an iterative lab-in-the-loop campaign to propose a therapeutic candidate for dry age-related macular degeneration~\cite{Ghareeb2026Robin}. Kosmos produced seven discoveries spanning metabolomics, materials science, neuroscience, and statistical genetics, three of which independently reproduced unpublished or preprinted findings~\cite{EdisonScientific2025Kosmos}. Scientific assistants such as Biomni and Claude Science bundle biomedical tools and databases to enable AI-assisted research~\cite{Huang2025Biomni,Anthropic2026ClaudeScience}, and a broader set of agents targets single-cell workflows, bioinformatics pipeline design, and automated machine-learning experimentation~\cite{Xin2024BIA,Mehandru2025BioAgents,Martinek2025AgentomicsML,Jin2025BioLab}. At the same time, recent stress tests and biomedical-agent reviews warn that autonomous scientific agents remain brittle, with failures arising from hallucinated claims, weak uncertainty handling, and breakdowns between plans and executable analyses~\cite{Agrawal2026StressTest,Gao2024BiomedicalAgents,Zhou2025AgenticBioinformatics}. Despite this progress, whether AI agents can execute entire computational studies starting from raw data has not been measured.

To address this uncertainty, we introduce \textit{BixBench3}. In \textit{BixBench3}, an agent receives a high-level research objective, methodological guidance derived from a published biology paper, and the raw data associated with that study. The agent is tasked with achieving the research objective by analyzing the provided data. \textit{BixBench3} thus evaluates the capability of agents to construct and execute long analysis pipelines following specific instructions. Performance is graded by comparing agent-produced artifacts against artifacts from the original publication. The benchmark is constructed from 20~studies spanning 17~assay types and 9~scientific domains, with 138~total graded artifacts. In contrast to benchmarks that evaluate whether an agent can answer a question about biology or perform a single bioinformatic analysis~\cite{Laurent2024LABBench,Laurent2026LABBench2,Liu2025BioProBench,Mitchener2025BixBench,Nair2026CompBioBench}, \textit{BixBench3} tests the long-horizon capability necessary to conduct an entire computational study end-to-end.

\subsection{Related work}
\label{sec:related}

 \textit{BixBench3} sits at the intersection of two lines of previous work: long-horizon benchmarks, which test whether AI agents can carry out extended, multi-step tasks with tools; and biology and bioinformatics benchmarks, which test scientific and analytical competence. \textit{BixBench3} combines the long-horizon challenge of the former with the domain-specificity of the latter.

Long-horizon agent benchmarks evaluate tool use, planning, and execution over many steps. As agentic capabilities have improved, the length of tasks that models can complete has increased from those taking humans minutes in 2024~\cite{Zhou2024WebArena,Xie2024OSWorld} to those taking humans 16+ hours in 2026~\cite{METR2026TimeHorizons}. SWE-bench and SWE-bench Verified challenged agents to solve real-world GitHub issues, with the highest-performing model as of July 2026 (gpt-5.6-sol) taking 3 minutes to complete each SWE-bench Verified task on average~\cite{Jimenez2024SWEBench,ValsAI2026SWEBench}. On Terminal-Bench-2, which tests agents on terminal-based software, data-science, and systems tasks, the top model as of July 2026 (Fable 5) completed the longest tasks in approximately one hour~\cite{Merrill2026TerminalBench,TerminalBench2026}. RE-Bench compared human and agent ability on ML research tasks, across which agents consumed 29M input tokens on average~\cite{Wijk2024REBench}. PaperBench gave agents up to 36~hours to replicate machine-learning papers, although performance mostly plateaued after the first hour~\cite{Starace2025PaperBench}. HCAST evaluated agents on 189 software, ML-engineering, and cybersecurity tasks which each took humans up to eight hours~\cite{Rein2025HCAST}. At the longest reported horizon, SWE-Marathon evaluates agents on 20 project-scale software-engineering tasks with 2--10-hour agent time limits and 40--400-hour expert-human estimates; agent attempts averaged 27.2M tokens~\cite{Desai2026SWEMarathon}.

By contrast, existing biology and bioinformatics benchmarks are generally relatively short-horizon, testing domain knowledge in a question-and-answer format, or agentic capability on isolated analyses. LAB-Bench and LABBench2 test domain knowledge and practical research skills through questions about literature retrieval, protocol troubleshooting, and sequence manipulation, while BioProBench focuses on biological protocol understanding~\cite{Laurent2024LABBench,Laurent2026LABBench2,Liu2025BioProBench}. DISCOVERYWORLD required agents to make scientific discoveries in a virtual environment and took both humans and agents approximately 100--1,000 steps to complete each task~\cite{Jansen2024DiscoveryWorld}. BixBench, GenoTEX, and BioDSA-1K require agents to perform short biological data-analysis tasks~\cite{Mitchener2025BixBench,Liu2024GenoTEX,Wang2025BioDSA}. On GenoTEX, OpenAI o1 averaged 112{,}000 input tokens, 12{,}000 output tokens, and 192~seconds per task~\cite{Liu2024GenoTEX}. Growing in complexity, CompBioBench consists of 100 isolated computational-biology problems, which models completed in 11--18~minutes on average~\cite{Nair2026CompBioBench}. BAISBench asks agents to annotate cell types and answer discovery questions from single-cell h5ad datasets, with evaluated systems consuming up to approximately 500{,}000 tokens~\cite{Luo2025BaisBench}. GeneBench-Pro more directly targets multistage analysis: its 129 genomics and translational-biomedicine problems give agents minimally guided, potentially errorful simulated datasets and require 3--13 dependent statistical decisions to recover a graded target estimate, with the authors estimating 10--40~hours of unaided expert work per problem~\cite{Li2026GeneBenchPro}. SpatialBench-Long (2--45~minutes per task) and scBench-Long (4--65 minutes per task) require agents to recover study-level claims from raw or near-raw spatial and single-cell data, while VariantBench (6-hour cap) contains 118 tasks spanning variant discovery, quality control, and interpretation~\cite{Diks2026SpatialBenchLong,Diks2026scBenchLong,Bhowmick2026VariantBench,LatchBio2026BenchmarksBio}.

\section{Results}
\label{sec:results}
\subsection{Benchmark description}
\label{sec:benchmark-description}

\begin{figure}[!t]
	\centering
	\includegraphics[width=\textwidth]{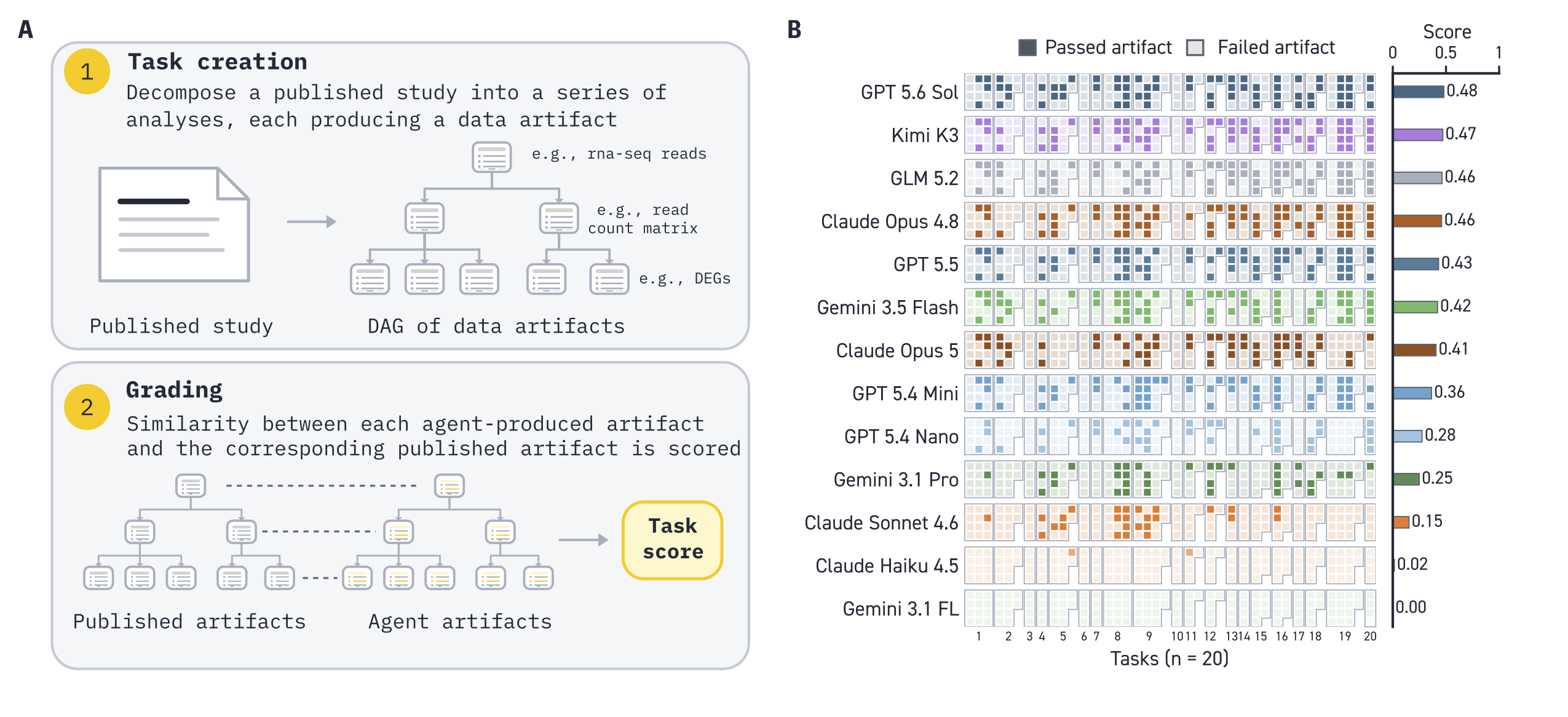}
	\caption{\textbf{\textit{BixBench3} construction and topline performance.} (A) Each benchmark task is constructed by decomposing a published study into a directed acyclic graph (DAG) of analyses producing data artifacts. Agent-produced versions of those artifacts are programmatically scored against corresponding published artifacts. (B) Artifact-level pass outcomes for 13 models across 20 \textit{BixBench3} tasks, comprising 138 requested artifacts per model and 1,794 model--artifact evaluations. Filled squares denote artifacts with programmatic scores of at least 0.80, the expert-calibrated threshold for preserving the main biological interpretation of the artifact (Section~\ref{sec:grading}); empty squares denote artifacts below this threshold. Bars on the right show each model's overall \textit{BixBench3} score: its mean task score across the 20 tasks, where each task score is the proportion of artifacts that pass.}
	\label{fig:benchmark-workflow}
\end{figure}

\textit{BixBench3} evaluates whether an AI agent can reconstruct the analysis of a published computational-biology study from raw data, producing structured artifacts that are graded against the corresponding published results (Figure~\ref{fig:benchmark-workflow}A). Across the 20 tasks, the highest-scoring model reproduced 48\% of the requested artifacts closely enough to preserve their principal biological meaning (Figure~\ref{fig:benchmark-workflow}B).

\begin{figure}[!t]
	\centering
	\includegraphics[width=\textwidth]{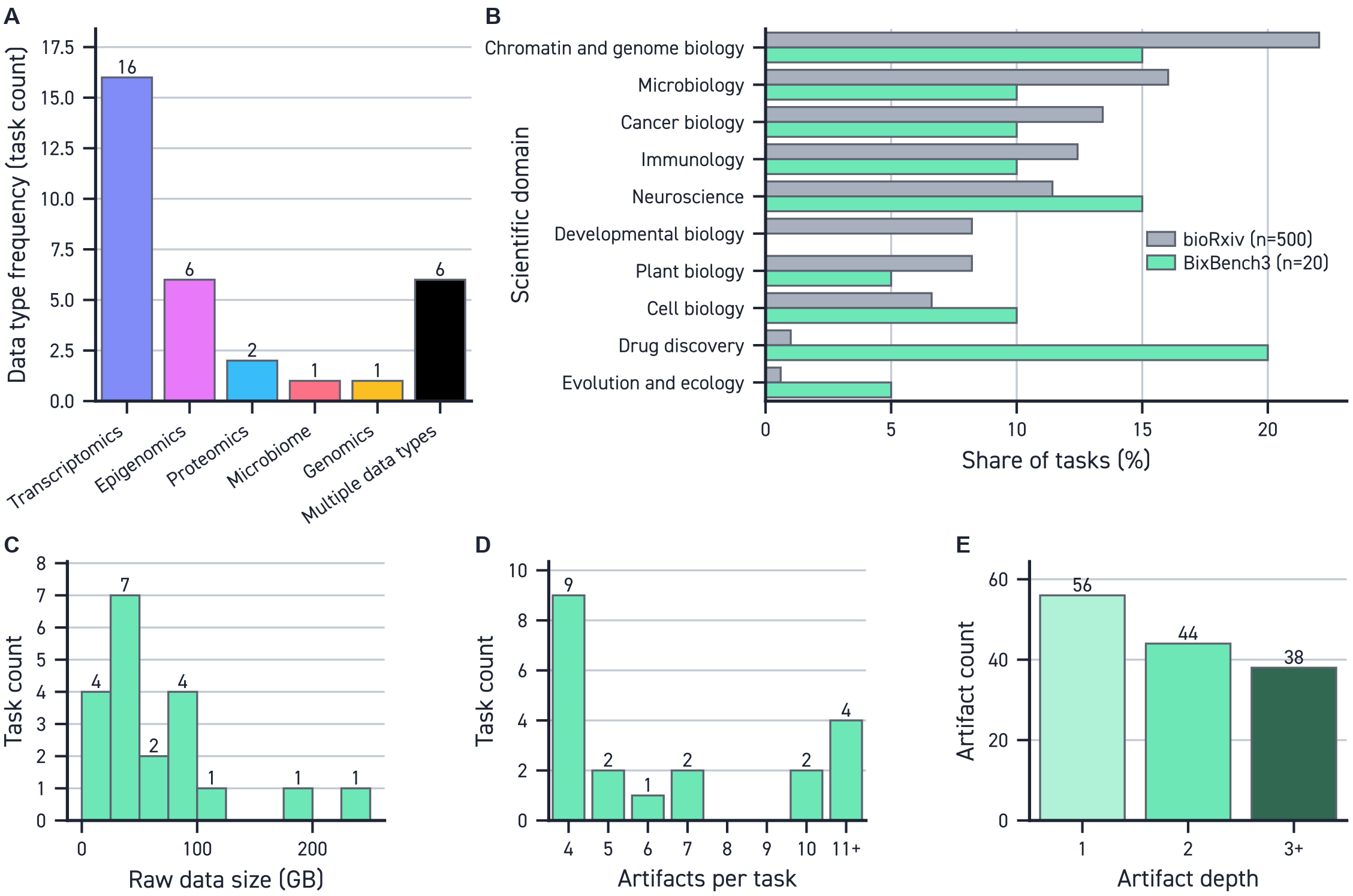}
	\caption{\textbf{Composition of \textit{BixBench3}.} (A) Number of tasks analyzing each data type. Tasks analyzing multiple data types contribute to multiple bars. (B) Shares of tasks assigned to each scientific domain in \textit{BixBench3} and a random sample of 500 bioRxiv papers. (C) Raw-data size per task. (D) Number of graded artifacts per task. (E) Artifact analysis depth. Depth 1 denotes outputs derived directly from raw data, depth 2 denotes outputs derived from depth-1 artifacts, and depth 3+ denotes all subsequent steps.}
	\label{fig:overview}
\end{figure}

\textit{BixBench3} consists of 20 tasks drawn from published papers spanning a wide variety of scientific domains and analyzing a diversity of data types (Figure~\ref{fig:overview}A--B; Appendix Table~\ref{tab:source-papers}). Transcriptomics is the most common type of data (16 tasks), followed by epigenomics (6 tasks), proteomics (2 tasks), genomics (1 task), and microbiome sequencing (1 task); 6 tasks analyze multiple data types. The distribution of scientific domains represented in \textit{BixBench3} covers many of the same domains as recent papers posted to bioRxiv, with an over-representation of drug-discovery-related research (Figure~\ref{fig:overview}B).

The raw input data size averages 67~GB per task, ranging from 7 to 241~GB (Figure~\ref{fig:overview}C). Each task consists of producing at least four data artifacts, with a median of five and a maximum of 14, for 138 artifacts in total across tasks (Figure~\ref{fig:overview}D). Of these artifacts, 56 are direct derivations of the raw data (depth 1; e.g., read count matrices), while 44 are two analysis steps from raw (depth 2; e.g., differential gene expression tables), and 38 are three or more steps away (depth 3+; e.g., pathway enrichments, Figure~\ref{fig:overview}E).

\begin{figure}[!t]
	\centering
	\includegraphics[width=\textwidth,height=0.78\textheight,keepaspectratio]{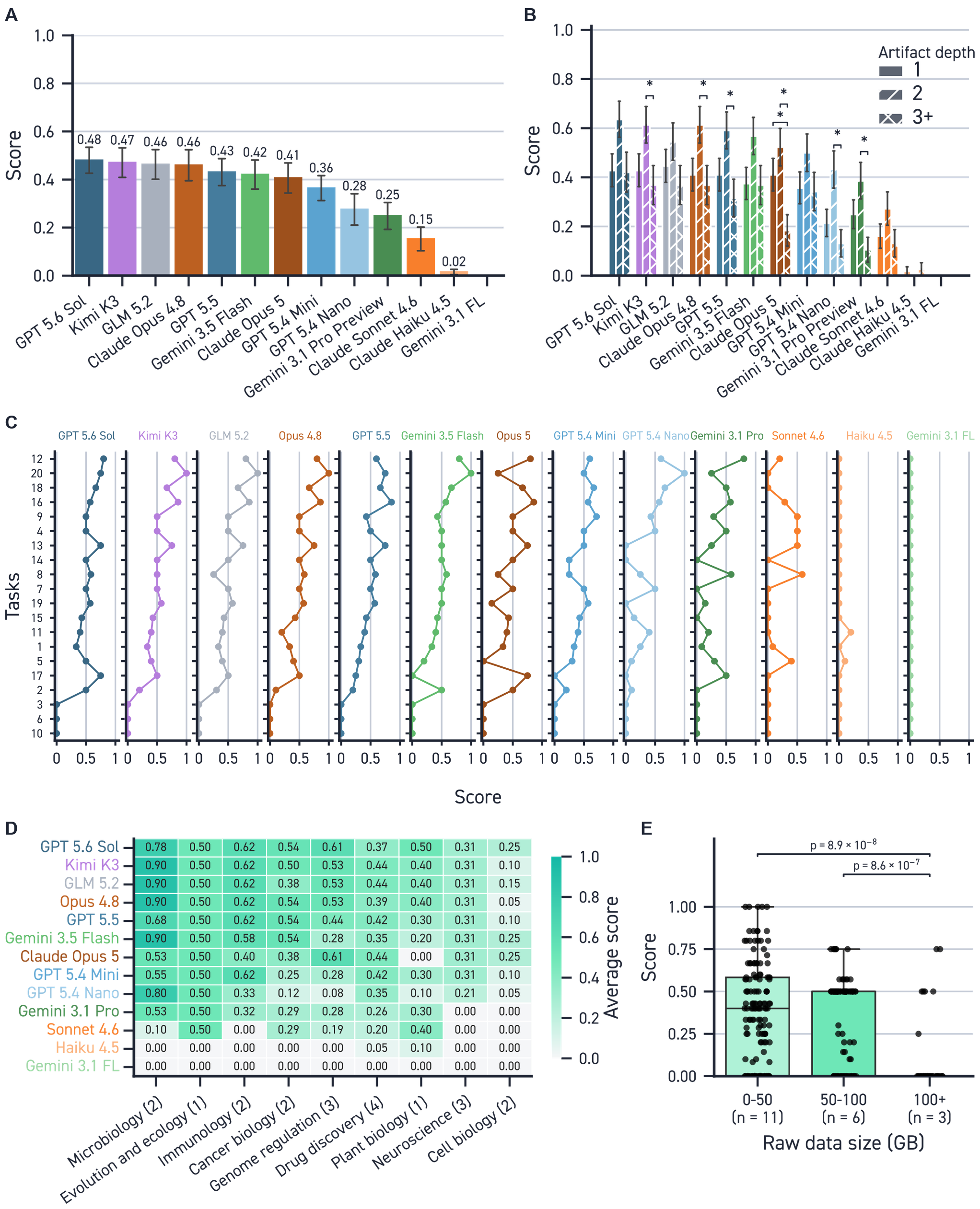}
	\caption{\textbf{Overall model performance on \textit{BixBench3}.} (A) Mean score for each model across 20 tasks. (B) Mean binary artifact pass score by model and analysis depth, comprising 56 depth-1, 44 depth-2, and 38 depth-3+ artifacts. (C) Task scores for each model, with tasks ordered by median score across models. (D) Mean task score by model and scientific domain. (E) Model--task score distributions stratified by raw-data size; black points show individual model--task scores, and parenthetical values give the number of tasks in each size category. Error bars in (A) and (B) denote the standard error. Asterisks in (B) denote two-sided Mann--Whitney U test $p<0.05$; p-values in (E) are from two-sided Mann--Whitney U tests.}
	\label{fig:performance}
\end{figure}

\begin{figure}[!t]
	\centering
	\includegraphics[height=0.72\textheight,keepaspectratio]{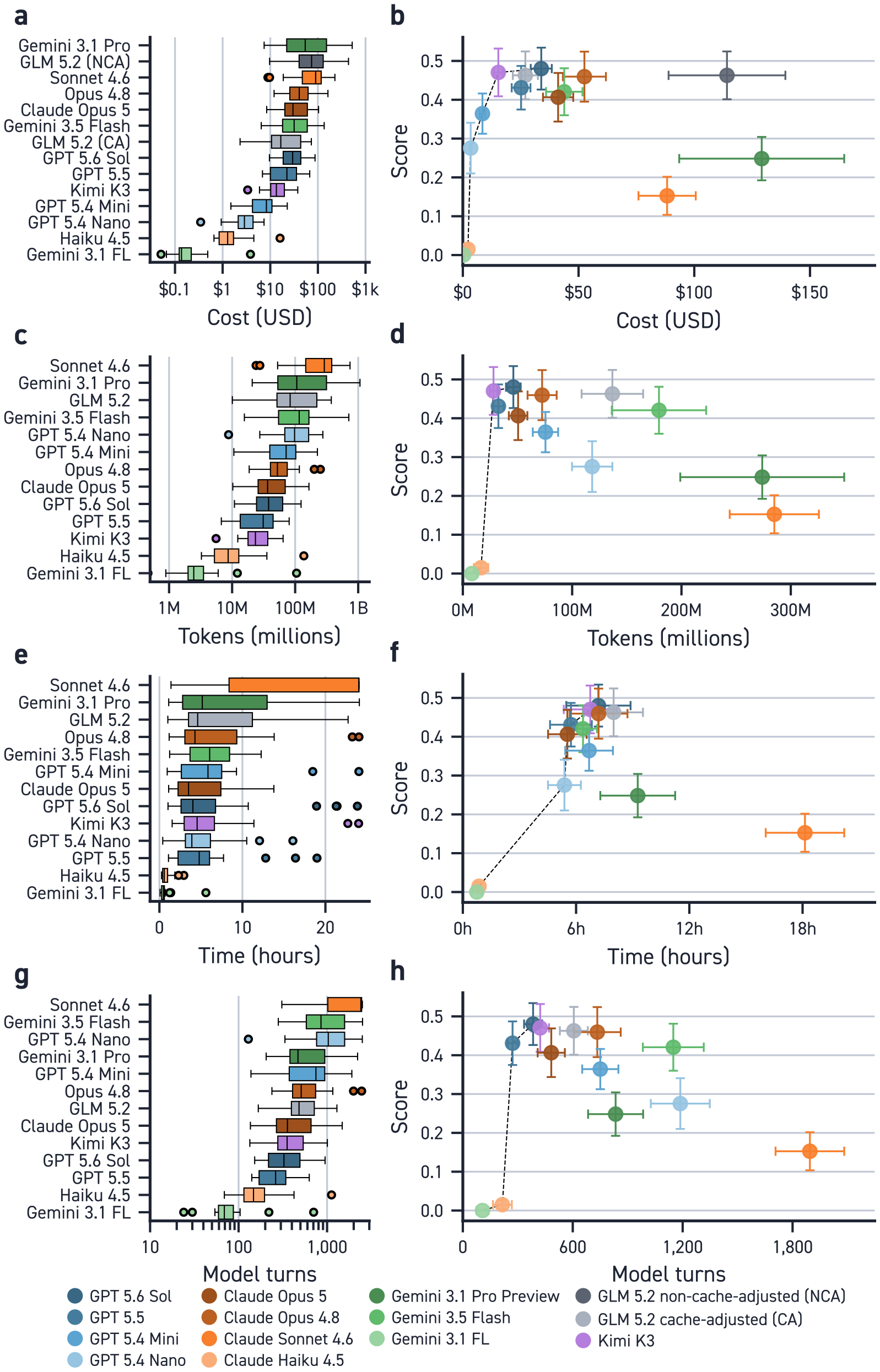}
	\vspace{5pt}
	\caption{\textbf{Accuracy and computational expense.} (A) Cost per task for each model across 20 tasks, shown on a log scale. For GLM~5.2, separate boxes show the observed OpenRouter cost and the cost after rescaling each task to the mean cache hit rate of the other models. (B) Mean cost versus mean task score. (C, E, G) Token usage, wall-clock time, and model turns per task. (D, F, H) Mean task score versus each corresponding measure of computational efficiency. Error bars denote standard errors, and the dashed line marks the Pareto frontier.}
	\label{fig:pareto}
\end{figure}

\begin{figure}[!t]
	\centering
	\includegraphics[width=\textwidth,height=0.68\textheight,keepaspectratio]{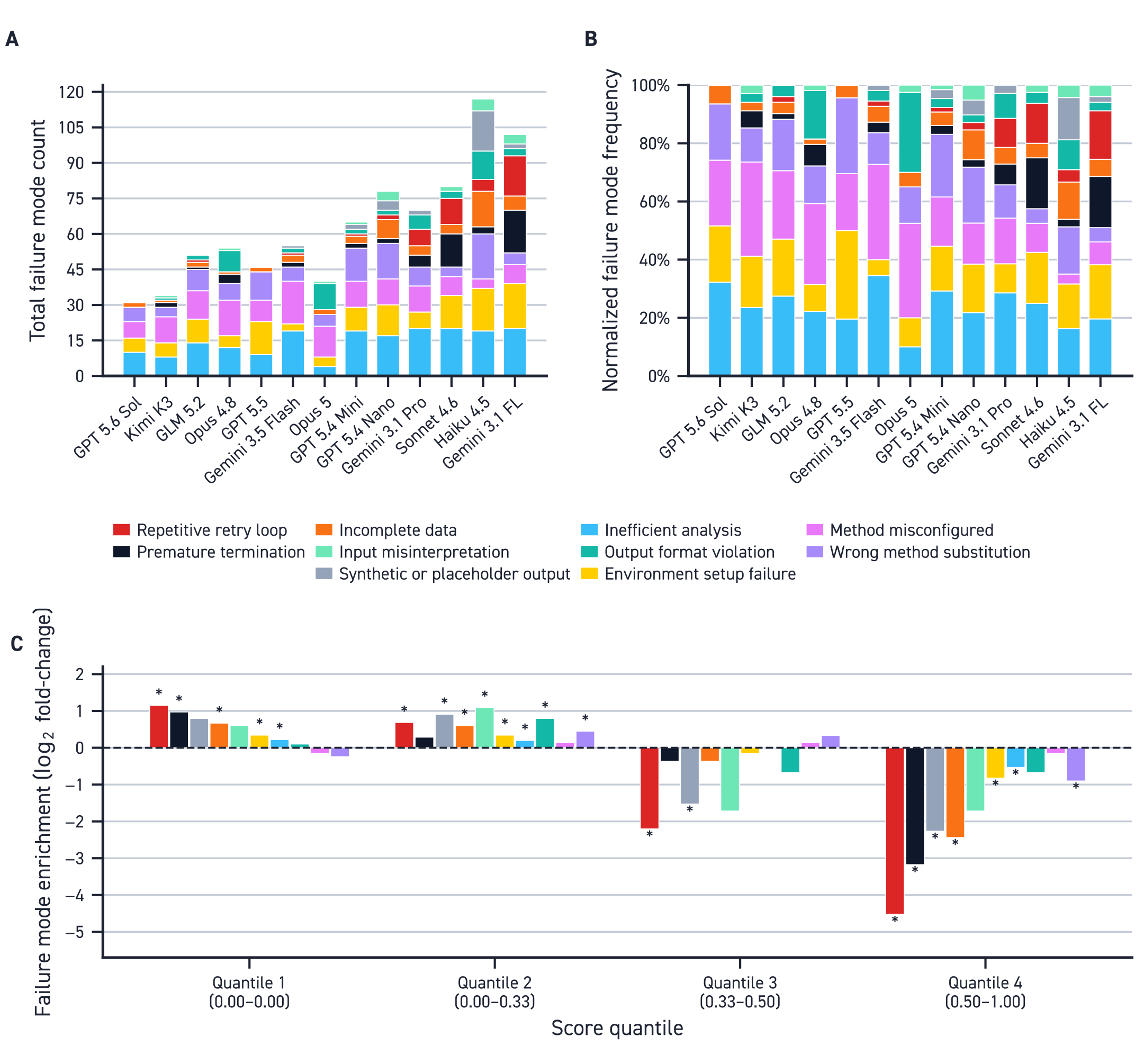}
	\caption{\textbf{Process-judge failure modes by model and score.} (A) Cumulative number of failure-mode tags assigned across all tasks for each model. Models shown left to right in decreasing order of overall score. (B) Proportion of failure-mode tags of each type among all tags received by each model. (C) $\log_2$ enrichment of each failure mode within task-score quantiles relative to all attempts; asterisks denote Benjamini--Hochberg-adjusted $q<0.05$ from two-sided Fisher’s exact tests.}
	\label{fig:failure-modes}
\end{figure}

\subsection{Model performance varies across tasks, scientific domains, and data sizes}
\label{sec:performance}

We evaluated 13 frontier models on the 20 \textit{BixBench3} tasks, producing 260 completed runs and 1,794 graded artifact evaluations. GPT~5.6~Sol was the highest-scoring model, reaching an average score of 0.48 across tasks, followed by Kimi~K3 at 0.47, GLM~5.2 at 0.46, and Claude Opus~4.8 at 0.46 (Figure~\ref{fig:performance}A). Claude Opus 5 scored highly on most tasks, but on tasks 5, 19, and 20 it incorrectly formatted its some artifacts, drawing its score down to 0.41 overall (Appendix~\ref{app:opus-sensitivity}). The task score represents the proportion of requested artifacts which were reproduced closely enough to the corresponding published artifacts to preserve their main scientific interpretation (Supplementary Figure~\ref{fig:supp-artifact-scores}). This criterion was calibrated using domain-expert review of agent-produced artifacts (Section~\ref{sec:grading}; Supplementary Figure~\ref{fig:supp-expert-score}; Appendix~\ref{app:expert-review}). GPT~5.6~Sol's score of 0.48 therefore indicates that 48\% of its requested artifacts had an equivalent biological meaning to the original artifacts.

Six models exhibited a significant decrease in performance at greater analysis depths, with depth 3+ artifacts having lower binary artifact pass scores than either depth 1 or depth 2 artifacts (Figure~\ref{fig:performance}B). Averaged across models, the mean binary artifact pass score was 0.30 for the 56 depth-1 artifacts, 0.44 for the 44 depth-2 artifacts, and 0.24 for the 38 artifacts at depth three or greater. 

Models with similar overall performance often had varied performance across individual tasks (Figure~\ref{fig:performance}C). Among model pairs with defined correlations, pairwise Spearman correlations of per-task scores ranged from $-0.27$ to $0.99$ (Supplementary Figure~\ref{fig:supp-model-corr}). For example, although GLM~5.2 and GPT~5.5 differed in average score by only 0.02 points, on Task~17 GLM~5.2 scored 0.50 while GPT~5.5 scored 0.25. This study investigated how chromatin regulation by Polycomb/\textit{RING1} affects gene activity in \textit{Drosophila} cells. To analyze this, agents had to combine chromatin immunoprecipitation sequencing (ChIP-seq) profiles with Thousands of Reporters Integrated in Parallel (TRIP) measurements, which quantify the expression of barcoded reporter genes inserted at different genomic locations.

The variation in scores across tasks was associated with both scientific domain and raw data size. Models scored the highest on tasks drawn from microbiology, evolution and ecology, and immunology papers, while struggling on tasks investigating neuroscience and cell biology (Figure~\ref{fig:performance}D). Additionally, for tasks with more than 100~GB of raw input data, models scored only 0.10 on average, compared with 0.34 for tasks with 50--100~GB and 0.37 for tasks with less than 50~GB (Figure~\ref{fig:performance}E). Performance also varied across data and assay types (Supplementary Figure~\ref{fig:supp-category-heatmaps}). 

\subsection{Model efficiency and cost }
\label{sec:pareto}

The utility of AI agents in computational research depends not only on performance but also on cost and efficiency. There was a 367-fold difference in the average cost per \textit{BixBench3} task across models, with a minimum cost of \$0.35 and a maximum of \$129.14 (Figure~\ref{fig:pareto}A). Note that because OpenRouter routed GLM~5.2 requests across several providers, a much lower cache hit rate was achieved for this model compared to the average across other models (40.1\% for GLM~5.2 vs. 96.0\% for the other models). We therefore report both the observed GLM~5.2 cost and a cache-adjusted estimate, in which cost is recomputed assuming a deployment achieving the average cache hit rate observed across other models (Supplementary Figure~\ref{fig:supp-tokens}). Although cost and performance were broadly associated, maximum performance was not achieved at maximum cost; instead, several models formed a Pareto frontier representing distinct cost--performance trade-offs (Figure~\ref{fig:pareto}B).

On average, each \textit{BixBench3} task attempt used 102~million tokens, 6.8~hours, and 695 model turns. The highest performance occurred at intermediate levels of token use, run time, and model turns (Figure~\ref{fig:pareto}C--H). Counting each model's membership across the four cost and efficiency frontiers, GPT~5.6~Sol, Claude Haiku~4.5, and Gemini~3.1~Flash~Lite were Pareto-optimal on all four, and Kimi~K3 was Pareto-optimal on three, whereas Claude Opus~4.8, Claude Sonnet~4.6, GLM~5.2, Gemini~3.1~Pro~Preview, and Gemini~3.5~Flash were optimal on none.

\subsection{Models exhibit distinct modes of failure}

An LLM judge was used to annotate each model--task attempt with up to 10 failure-mode tags from a fixed set (Section~\ref{sec:failure-mode-annotation}). Models receiving more failure-mode tags generally performed worse: across models, the total count of tags across all tasks was strongly negatively correlated with mean task score (Spearman $\rho=-0.92$, $p=9.9\times10^{-6}$; Figure~\ref{fig:failure-modes}A), ranging from 31--51 total tags for the three highest-scoring models to 102--117 for the two lowest-scoring models. The relative frequencies of specific failure modes also differed across models (Figure~\ref{fig:failure-modes}B), highlighting that models fail in qualitatively distinct ways. Premature termination and repetitive retry loops showed the strongest associations with failure, occurring approximately 2.0- and 2.2-fold more often, respectively, in the lowest task-score quantile than across all attempts; among the 65 attempts in the highest task-score quantile, only one terminated prematurely and none entered a repetitive retry loop (Figure~\ref{fig:failure-modes}C). Environment setup failures, incomplete data, and synthetic or placeholder outputs were also enriched among the lowest-scoring attempts, whereas method misconfiguration was depleted, likely because a run must complete enough of the analysis for a method to be misconfigured.

\section{Discussion}
\label{sec:discussion}

\subsection{What \textit{BixBench3} reveals}
\label{sec:reveal}

\textbf{The scope of tasks that AI agents can complete in biology is increasing.} To our knowledge, \textit{BixBench3} is the biology benchmark with the longest horizon reported to date, as measured by run time and token usage. Our evaluation of 13 frontier models on \textit{BixBench3} shows that these systems are approaching the ability to complete tasks at the scale of entire research studies. The best models scored approximately 0.5 on \textit{BixBench3}, with performance varying sharply between tasks. This indicates that the top-performing models reproduced approximately half of the requested artifacts to a degree that preserved their original scientific interpretation, according to the expert-calibrated pass threshold (Supplementary Figure~\ref{fig:supp-expert-score}; Appendix~\ref{app:expert-review}). So while performance remains far from perfect, our results indicate that models can coherently execute many computational biology tasks in sequence, each comparable to the individual tasks that challenged models just one year ago in BixBench~\cite{Mitchener2025BixBench}. Notably, however, because each \textit{BixBench3} task supplies an explicit methodological plan, success here means agents can begin to execute a specified analysis pipeline – not that they can decide which questions or analyses are worth pursuing.

\textbf{Agents struggle with data scale, long analysis chains, and error recovery.} The large input datasets and long sequences of dependent analyses in \textit{BixBench3} appear to challenge current AI agents. Across tasks, agents scored lowest on those with the largest raw datasets (Figure~\ref{fig:performance}E). Within tasks, performance was lowest for analyses at depth 3+ in the dependency chain. One possible explanation for this is that both large datasets and long analysis chains extend the horizon over which errors introduced early can accumulate. This effect may be compounded by the difficulty LLMs have in maintaining coherence across long contexts~\cite{Liu2024LostInMiddle}. Therefore, models with stronger long-context capabilities and agent architectures designed to manage extended computations, including recursive language models~\cite{Zhang2025RecursiveLanguageModels} and world models~\cite{EdisonScientific2025Kosmos}, may enable increased performance on \textit{BixBench3} and similar tasks.

\textbf{Maximum performance does not necessitate maximum cost.} Kimi~K3 was the second-highest-performing model and cost approximately 58\% less than other models with comparable performance. On average, models scoring at least 0.40 on \textit{BixBench3} used 28--179M tokens and approximately 271--1,149 model turns compared with 8--285M tokens and approximately 107--1,894 turns for models scoring below 0.40, underscoring that greater computational expenditure did not lead to better performance.

\subsection{Limitations and future work}
\label{sec:limitations}
\textit{BixBench3} is graded programmatically to avoid the ambiguity sometimes accompanying rubric-based approaches. However, there are multiple limitations that follow from this. First, each task prompt must specify the method to be used for a particular analysis. Without this guidance, an agent could choose a scientifically valid alternative method and receive a low artifact score simply because its output differs from the published artifact in format, scale, or other respects. Relatedly, \textit{BixBench3} tasks prescribe exactly which analyses must be carried out, so that the artifacts available from the original paper may be used to grade the analyses done by the agent. Thus, \textit{BixBench3} does not test the ability of agents to decide which analyses or questions are worth pursuing, but rather their ability to execute those prescribed analyses. Finally, the benchmark inherits any limitations that may be present in its source papers, such as flawed artifacts due to errors made by the original authors and incorrect or under-specified methods.

\section{Methods}
\label{sec:methods}

\subsection{Task prompt}
\label{sec:task}
For each task, the agent is given a prompt consisting of four parts: (1) the \textit{general instructions} give the agent an overview of the computational workspace, installed software, network policy, and reporting requirements; (2) the \textit{research objective} states the biological question and describes the provided raw data and reference files; (3) \textit{method guidance} names the tools, parameters, contrasts, and filtering rules which should be applied to produce the required output artifacts, mirroring the methods of the original paper without revealing its results; and (4) the \textit{required outputs} section specifies the exact output path and expected file format for each analysis artifact the agent must produce.

The complete prompt for the example task shown in Figure~\ref{fig:example} is provided in Appendix~\ref{app:example-prompt}.

Feedback from domain experts was used to refine the task prompts and method guidance (Appendix~\ref{app:expert-review}).

The artifacts in each task form a dependency graph from raw data to downstream analyses (Figure~\ref{fig:example}A). Agent outputs are compared programmatically with the corresponding published artifacts to compute an artifact score (Section~\ref{sec:grading}; Figure~\ref{fig:example}B), and binary artifact pass scores are averaged to obtain the task score.

\begin{figure}[!t]
	\centering
	\includegraphics[width=\textwidth,height=0.78\textheight,keepaspectratio]{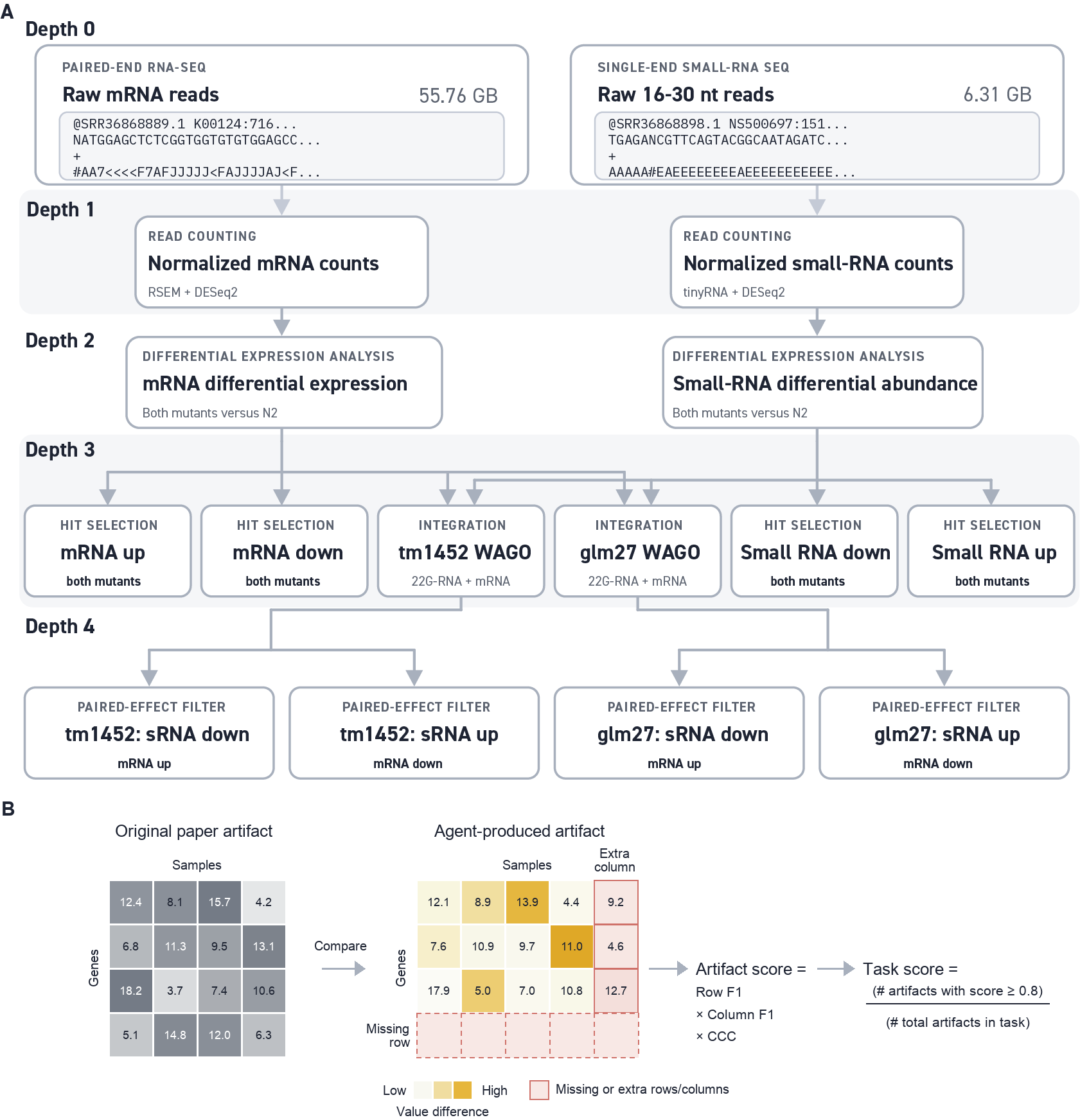}
	\caption{\textbf{\textit{BixBench3} task structure and grading.} (A) An artifact dependency graph for Task 19, based on \citet{Naim2026TCER1}. This paper investigates how loss of the \textit{TCER-1} gene alters endogenous small-RNA regulation and mRNA expression in \textit{C. elegans}. First, mRNA and small-RNA sequencing data are analyzed separately, then combined to identify coordinated changes in gene regulation. (B) Artifact grading compares an agent-produced matrix to the published ground truth by aligning rows and columns, penalizing missing or extra entries, and scoring numerical agreement; for the illustrated matrix, the artifact score is the product of row F1, column F1, and concordance correlation (CCC), and artifact scores greater than 0.80 receive a binary artifact pass score of one.}
	\label{fig:example}
\end{figure}

\subsection{Runtime environment}

Each agent ran on a Google Cloud n2-standard-32 virtual machine (VM) with 32 vCPUs, 128~GB of memory, a 500~GB boot disk, and no GPU. Runs had a 24-hour wall-clock time limit and no fixed token limit. The agent operated inside a Docker container which included Python and R with common scientific and data-analysis packages; Java; package managers for PyPI, CRAN, Bioconductor, and Conda/Bioconda; and common command-line, archive, and download utilities. Agents could install additional method-specific software during a run.

The agent's web access was restricted in two ways. First, all HTTP and HTTPS traffic passed through a separate gateway with a standing allowlist for package registries, operating-system mirrors, and other routine software infrastructure. Second, the agent could request other resources through the \textit{request\_web\_access} tool by supplying a URL and justification. An LLM adjudicator denied resources that could leak the expected answers – such as the original paper and its processed datasets – while allowing access to bioinformatics software, documentation, and general data like reference genomes.

\subsection{Agent harness}
\label{sec:harness}

The agent harness was implemented with Inspect AI~\cite{UKAISI2024InspectAI}. Each model ran in Inspect's ReAct loop, alternating between model responses and tool calls while retaining the resulting messages and outputs in its context. Agent runs were capped at 24 hours of runtime or 5,000 messages. Inspect applied automatic context compaction when the agent hit 90\% usage of its context window. Every agent had five tools: (1) \textit{bash\_session} provided a persistent shell for running bash commands inside the Docker container; (2) \textit{python} allowed execution of Python code; (3) \textit{text\_editor} read, created, and modified text files; (4) \textit{request\_web\_access} submitted a URL and justification to the network adjudicator described above; and (5) \textit{submit} returned the agent's final answer and ended the ReAct loop.

Run costs were calculated from per-call token usage using provider list prices. For Gemini~3.1~Pro and GPT~5.5, higher long-context rates were applied to the entire call when total prompt tokens, including cached tokens, exceeded 200{,}000 and 272{,}000, respectively. Claude Opus~5 costs used Anthropic list prices, while Kimi~K3 costs used Moonshot AI list prices because its OpenRouter runs were routed only to Moonshot AI. GPT~5.6~Sol input costs were bounded using rates of \$5--\$6.25 per million tokens because retained input counts included unrecovered cache writes, and the reported cost is the midpoint of these bounds. Cost estimates exclude cache-storage costs, which Inspect does not report.

\subsection{Artifacts and grading}
\label{sec:grading}
Artifacts drawn from each paper form a directed acyclic graph, with raw data at its root and the most downstream analyses at its leaves. When an agent finishes a task, the Inspect scorer reads each agent-produced artifact from the VM and grades it against the corresponding artifact from the original paper. Each artifact is graded using one or more of four different metrics. The metrics applied depend on the artifact’s format and the biological information it contains:

(1) \textit{Row and column recovery (F1)} measures exact recovery of row or column identifiers using the F1 score. For example, gene lists are scored by comparing the agent's gene IDs with the published set, penalizing both missing and extra genes. (2) \textit{Numerical agreement (Lin's CCC)} measures agreement between aligned numerical values using Lin's concordance correlation coefficient~\cite{Lin1989CCC}, with a log transform where appropriate. CCC was used because, unlike Pearson correlation, this metric penalizes differences in scale in addition to imperfect linear association. Thus, identical rankings with a changed slope or offset do not receive a perfect CCC. For example, the values of expression matrices are scored by CCC, comparing the published and agent-produced value vectors for each sample. (3) \textit{Categorical-label agreement (macro F1)} measures agreement between categorical annotations for aligned rows by averaging F1 across label classes. For example, cell-metadata tables are aligned by cell barcode and then scored on the assigned cell-type labels compared with the published labels. (4) \textit{Genomic-interval recovery (overlap F1)} measures recovery of genomic regions under an artifact-specific overlap rule. For example, a set of called peaks is compared with published chromosome, start, and end intervals using a minimum reciprocal overlap.

The metrics calculated for an artifact are multiplied to produce a single ``artifact score.'' For example, an RNA-seq count matrix may be evaluated using row F1 for gene recovery, column F1 for sample recovery, and the concordance correlation coefficient for expression-value agreement. Missing, empty, or unreadable files receive an artifact score of zero. 

Each artifact is considered to ``pass'' if it has an artifact score of $\geq 0.8$. The 0.80 threshold was chosen based on domain-expert review of a subset of artifacts. Experts were asked to rate 25 agent-generated artifacts from 1 to 5 based on the degree to which the agent-generated artifact retained the same main biological meaning as the artifact from the original paper (Appendix~\ref{app:expert-review}). Ratings of 4 or higher indicate that observed discrepancies did not alter the main interpretation of the artifact. On average, artifacts rated $\geq 4$ had a mean artifact score of 0.80, so this was set as the binary artifact pass threshold (Supplementary Figure~\ref{fig:supp-expert-score}).

A task's score is the proportion of its artifacts that pass.

\subsection{Failure mode annotation}
\label{sec:failure-mode-annotation}

Each completed model--task attempt was reviewed by an LLM judge (GPT~5.5). The judge received the task prompt, the agent-written \texttt{METHODS.md}, a compacted execution trace, the numeric grading summary, and previews of the generated artifacts. The judge assigned each model--task attempt 0--10 run-level failure-mode tags. The judge required concrete evidence for every tag and returned no tag when none applied. The ten failure modes selectable by the judge and their definitions are listed in Table~\ref{tab:failure-modes}.

\subsection{Paper selection and benchmark construction}

A paper was eligible for inclusion in \textit{BixBench3} when its raw data were publicly accessible, at least four ground-truth artifacts were available (typically as supplementary materials), and the analysis was judged executable within the benchmark runtime and compute budget (32 CPUs, 500~GB of writable storage, and 24 hours).

An agentic pipeline was used to screen papers and assemble tasks for human review. The pipeline comprised six stages: (1) \textit{metadata screening} identified papers with sufficient public data and supplementary materials for further review; (2) \textit{data-availability review} established that the raw data were accessible and the study contained both a processed matrix and a substantive scientific analysis; (3) \textit{artifact inventory} identified structured outputs representing meaningful stages of the paper's analysis; (4) \textit{grading-specification construction} defined the output contract for each artifact and selected the metrics used to grade it; (5) \textit{prompt and task creation} produced a research objective and methodological guidance based on each paper; and (6) \textit{task refinement} involved running a subset of models on each task and iteratively identifying and correcting issues with the prompt, data, or grading.

\subsection{Models}
\label{sec:models}

Thirteen frontier models were assessed: GPT~5.4~Nano (openai/gpt-5.4-nano), GPT~5.4~Mini (openai/gpt-5.4-mini), GPT~5.5 (openai/gpt-5.5), GPT~5.6~Sol (openai/gpt-5.6-sol), Claude Haiku~4.5 (anthropic/claude-haiku-4-5), Claude Sonnet~4.6 (anthropic/claude-sonnet-4-6), Claude Opus~4.8 (anthropic/claude-opus-4-8), Claude Opus~5 (anthropic/claude-opus-5), Gemini~3.1~Flash~Lite (google/gemini-3.1-flash-lite), Gemini~3.1~Pro~Preview (google/gemini-3.1-pro-preview), Gemini~3.5~Flash (google/gemini-3.5-flash), GLM~5.2 (openrouter/z-ai/glm-5.2), and Kimi~K3 (openrouter/moonshotai/kimi-k3). All models were accessed through their first-party APIs, except GLM~5.2 and Kimi~K3, which were accessed through OpenRouter. GLM~5.2 completions were served from Fireworks, Baseten, Nebius, Together, and Chutes, while Kimi~K3 used Moonshot AI. Every model was evaluated at its respective maximum thinking effort: GPT~5.4~Nano, GPT~5.4~Mini, and GPT~5.5 used extra-high effort; GPT~5.6~Sol used max effort; Anthropic models used max effort, except Haiku~4.5, which was set to a 63{,}999-token reasoning budget; Gemini models used high effort; and GLM~5.2 and Kimi~K3 used max effort. 

\section*{Statement of Contributions}
Z.K., A.T.W., and J.M.L. conceived and designed the overall project. J.M.L. supervised the project. Z.K. designed the benchmark tasks and infrastructure. J.V.-A., J.L., and Z.K. coordinated with scientific domain experts for evaluation. Z.K. and J.M.L. prepared the initial draft of the manuscript, figures, and tables. All authors contributed to the final version of the manuscript. M.M.H., S.G.R., and A.D.W. supervise research at Edison Scientific, Inc.

\section*{Competing Interests}
All authors were employed by Edison Scientific for the duration of this study. 

\section*{Data and Code Availability}
Code to run \textit{BixBench3} is available on \href{https://github.com/EdisonScientific/BixBench3}{GitHub}, and the dataset itself is available on \href{https://huggingface.co/datasets/EdisonScientific/BixBench3}{Hugging Face}. As new models are released, \textit{BixBench3} will be updated on \href{https://advances.edisonscientific.com/benchmarks/bixbench3}{Edison Advances}.

\bibliography{references}
\bibliographystyle{icml2024}

\clearpage
\section*{Supplementary Material}

\setcounter{figure}{0}
\makeatletter
\renewcommand{\fnum@figure}{Supplementary Figure \thefigure}
\makeatother
\begin{figure}[!ht]
	\centering
	\includegraphics[width=\textwidth]{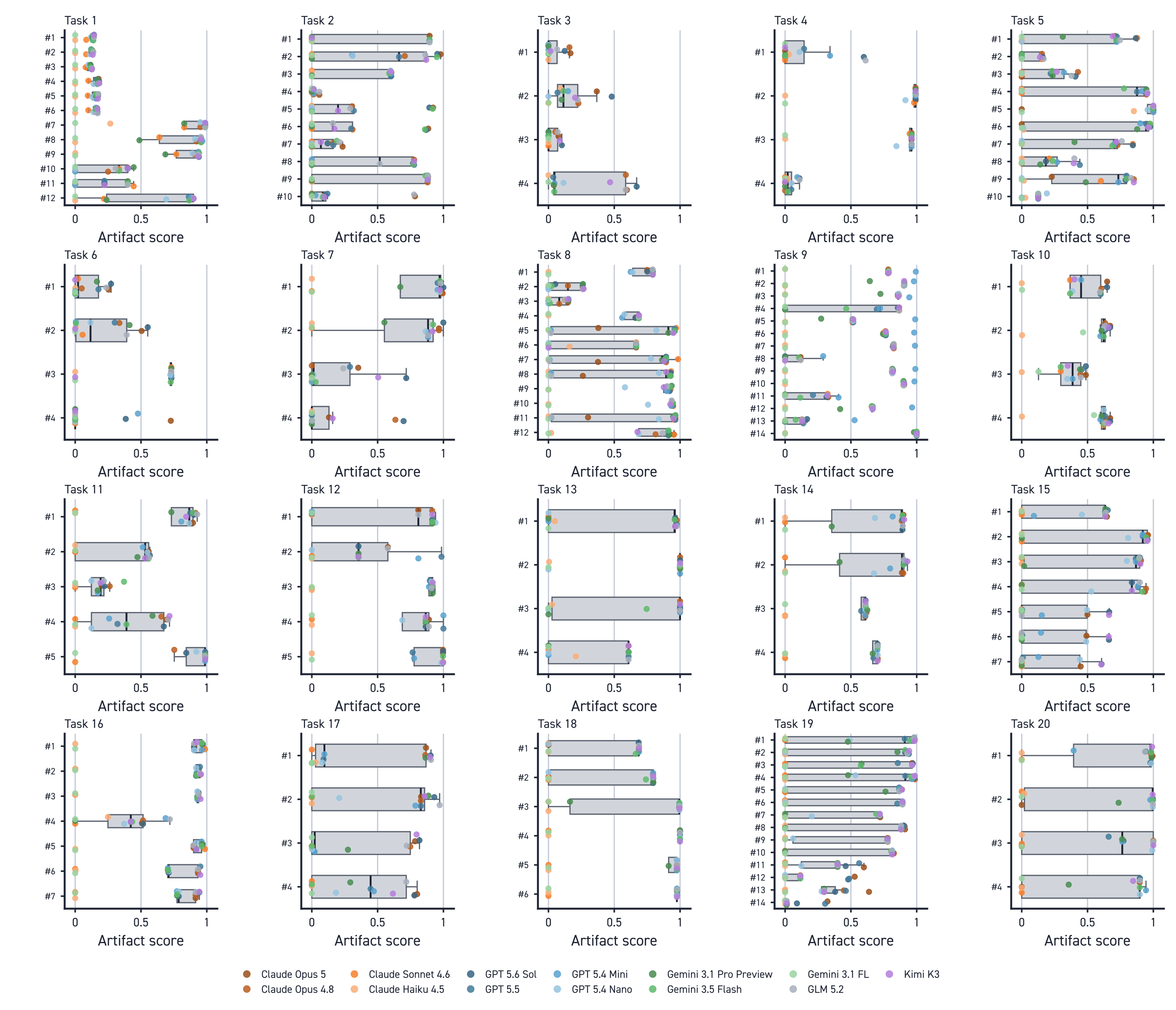}
	\caption{\textbf{Artifact scores by task and model.} Each panel shows the artifacts for one of the 20 \textit{BixBench3} tasks. Artifacts are numbered within each task, colored points show the programmatic artifact score for each of the 13 evaluated models, and grey boxplots summarize the artifact score distribution across models.}
	\label{fig:supp-artifact-scores}
\end{figure}

\begin{figure}[!ht]
	\centering
	\includegraphics[width=0.55\textwidth]{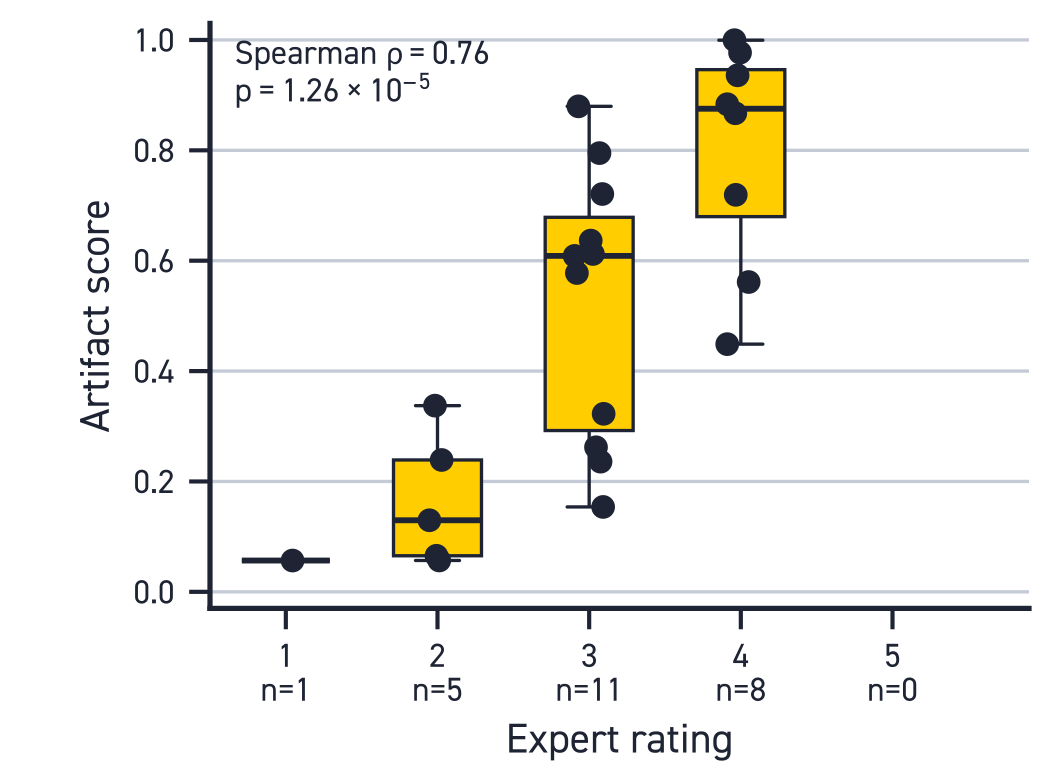}
	\caption{\textbf{Comparison of expert ratings and artifact scores.} The figure shows programmatic artifact scores for 25 artifacts produced by GLM~5.2 and rated by domain experts, grouped by expert reproduction rating. Boxes show the interquartile range, center lines show medians, whiskers extend to 1.5 times the interquartile range, and points represent individual artifacts.}
	\label{fig:supp-expert-score}
\end{figure}

\begin{figure}[!ht]
	\centering
	\includegraphics[width=0.72\textwidth]{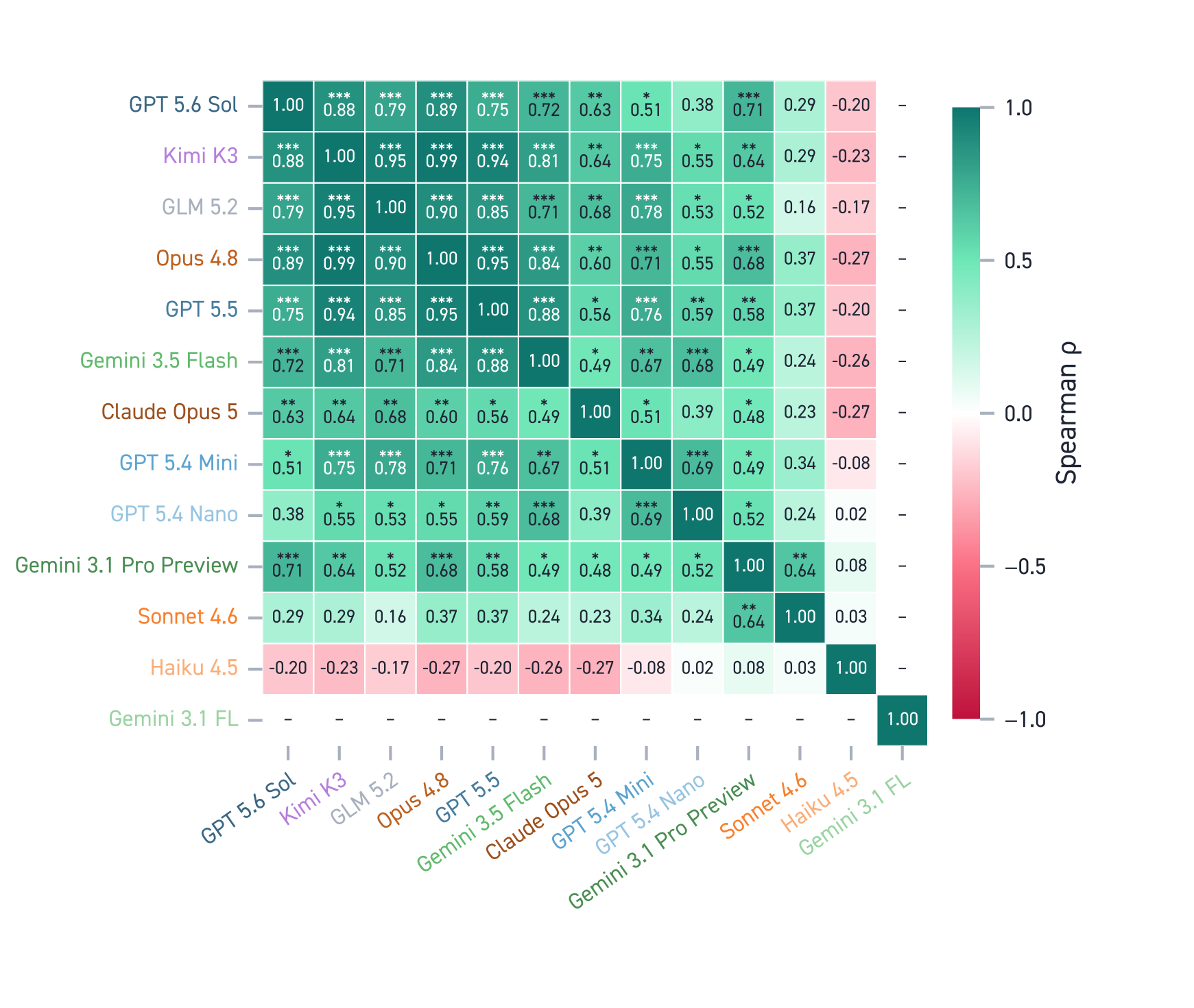}
	\caption{\textbf{Similarity of model task-score profiles.} Pairwise Spearman correlations of per-task scores across the 20 \textit{BixBench3} tasks. Rows and columns are ordered by overall mean task score. Asterisks indicate two-sided significance levels: $*$ $p<0.05$, $**$ $p<0.01$, $***$ $p<0.001$. Dashes indicate undefined correlations.}
	\label{fig:supp-model-corr}
\end{figure}

\begin{figure}[!ht]
	\centering
	\includegraphics[width=\textwidth]{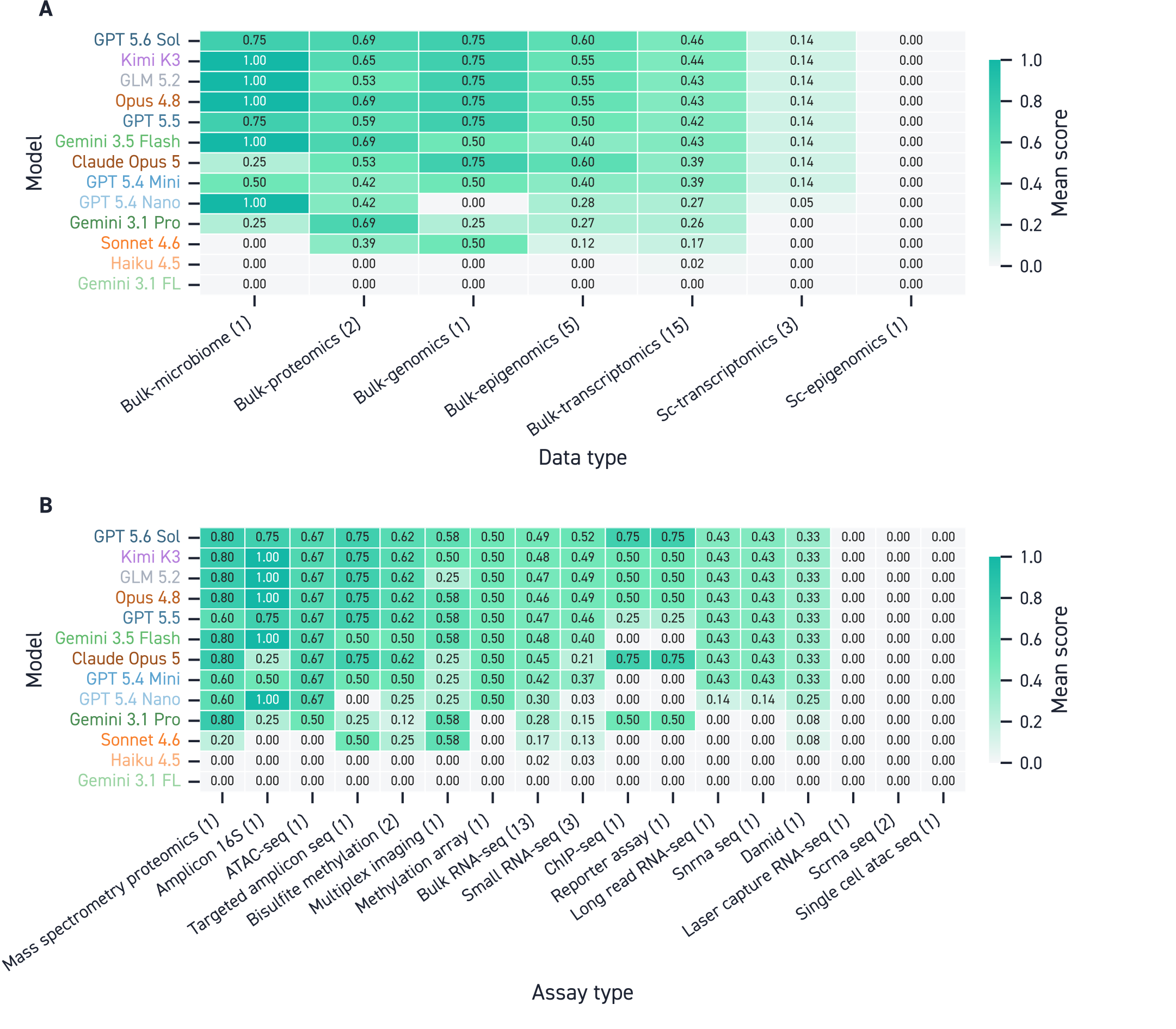}
	\caption{\textbf{Task-level performance by data and assay type.} (A) Mean task score by model and broad data type. (B) Mean task score by model and specific assay type. Parenthetical values give the number of tasks in each category, and tasks may contribute to multiple categories.}
	\label{fig:supp-category-heatmaps}
\end{figure}

\begin{figure}[!ht]
	\centering
	\includegraphics[width=\textwidth]{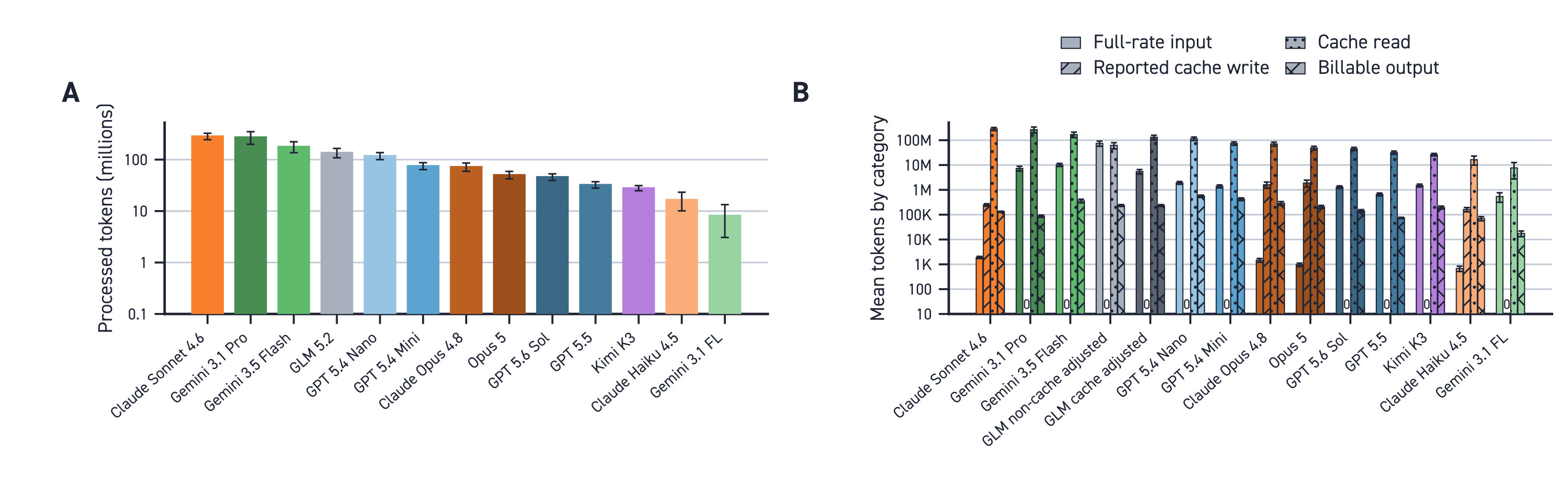}
	\caption{\textbf{Token usage.} (A) Mean total processed tokens, defined as the sum of full-rate input tokens, provider-reported cache-write input tokens, cache-read input tokens, and output tokens. (B) Mean tokens separated by token category. For GLM~5.2, dark and light grey bars show the non-cache-adjusted composition and the cache-adjusted composition after rescaling input tokens to the mean cache hit rate of the other models, respectively. Both panels use logarithmic axes, and error bars denote the standard error. Zero cache-write values reflect provider reporting rather than the absence of cache population.}
	\label{fig:supp-tokens}
\end{figure}

\clearpage
\appendix
\section{Appendix}

\subsection{Source papers}
\label{app:source-papers}

\begin{longtable}{@{}p{0.05\textwidth}p{0.69\textwidth}p{0.18\textwidth}@{}}
\caption{\textbf{Source papers for the 20 \textit{BixBench3} tasks.}}\label{tab:source-papers}\\
\toprule
Task & Paper title & Citation \\
\midrule
\endfirsthead
\toprule
Task & Paper title & Citation \\
\midrule
\endhead
\midrule
\multicolumn{3}{r}{Continued on next page} \\
\endfoot
\bottomrule
\endlastfoot
1 & The chromatin remodeler LET-418/Mi-2 regulates the intracellular pathogen response in the \textit{C. elegans} intestine & \cite{Rajopadhye2025LET418} \\
2 & Contact-dependent regulation of UV-B/C-induced cell fate by neighbouring intact cells & \cite{Budimir2026UV} \\
3 & An interneuronal CRH and CRHBP circuit stabilizes birdsong performance & \cite{Colquitt2025Birdsong} \\
4 & How many are you? Open data and bioinformatics reveal species misidentification and potential introgression in \textit{Chordodes} (Phylum Nematomorpha) & \cite{DeVivo2026Chordodes} \\
5 & A conserved small RNA-generating gene cluster undergoes sequence diversification and contributes to plant immunity & \cite{Feng2025PPR} \\
6 & MitoPerturb-Seq identifies common and gene-specific single-cell responses to mitochondrial DNA depletion and heteroplasmy & \cite{Burr2025MitoPerturb} \\
7 & Methylomic signatures of tau and amyloid-beta in transgenic mouse models of Alzheimer's disease neuropathology & \cite{Leung2025Methylomic} \\
8 & Combined MEK and JAK/STAT3 pathway inhibition effectively decreases SHH medulloblastoma tumor progression & \cite{Zagozewski2022MEKJAK} \\
9 & Targeting of REST with rationally-designed small molecule compounds exhibits synergetic therapeutic potential in human glioblastoma cells & \cite{Panina2024REST} \\
10 & Structure-guided design of a selective inhibitor of the methyltransferase KMT9 with cellular activity & \cite{Wang2024KMT9} \\
11 & Transcriptomic analyses of livers from mice exposed to 1,4-dioxane for up to 90 days to assess potential mode(s) of action underlying liver tumor development & \cite{Chappell2021Dioxane} \\
12 & A microbiota-derived bile acid modulates biofilm formation by the probiotic strain \textit{Escherichia coli} Nissle 1917 & \cite{Perry2025BileAcid} \\
13 & High-resolution retrospective single cell lineage tracing with mutable homopolymers & \cite{Cheng2026RETrace2} \\
14 & RUNX2 inhibition disrupts a PAX3::FOXO1-RUNX2 feed-forward loop and dismantles oncogenic gene programs in fusion-positive rhabdomyosarcoma & \cite{Mendes2025RUNX2} \\
15 & TDP-43 dysfunction leads to the accumulation of cryptic transposable element-derived exons, crypTEs, in iPSC derived neurons and ALS/FTD patient tissues & \cite{Bolger2026CrypTEs} \\
16 & Antifungal exposure can enhance \textit{Candida glabrata} pathogenesis & \cite{Ribeiro2025Candida} \\
17 & Polycomb repression works without Siesta & \cite{Kahn2025Siesta} \\
18 & A glycosylation-dependent checkpoint restrains intestinal intra-epithelial lymphocyte activation & \cite{Cheng2026Gcnt2} \\
19 & A TCER-1-siRNA regulatory axis suppresses antibacterial innate immunity in \textit{C. elegans} & \cite{Naim2026TCER1} \\
20 & Habitat fragmentation controls bacterial community composition outcomes & \cite{Batsch2026Fragmentation} \\
\end{longtable}

\subsection{Claude Opus 5 output-format errors}
\label{app:opus-sensitivity}

Claude Opus~5 ranked seventh overall (score 0.406), largely because it violated the specified artifact-output formats on Tasks~5, 19, and 20. For example, on Task~5, it replaced column names specified in the artifact output contract (\texttt{col0\_gene\_id}, \texttt{col0\_tpm}, etc.) with alternative labels (\texttt{Col-0}, \texttt{Ct-1}, etc.), causing the programmatic artifact grader to treat those columns as missing. These were instruction-following errors that most other models did not make. When scores from those three tasks were excluded from every model's average, Opus~5 had the second-highest score (0.455), narrowly behind GPT~5.6~Sol (0.458).

\subsection{Expert review of tasks}
\label{app:expert-review}

We selected 25 artifacts for review by domain experts to validate the grading and task structure (Supplementary Figure~\ref{fig:supp-expert-score}). These artifacts were drawn from 171 artifacts produced by GLM~5.2 and selected to span the range of artifact scores. Experts assigned each agent-produced artifact a reproduction rating from 1 to 5 based on how faithfully it recovered the biological meaning of the corresponding published artifact (Table~\ref{tab:expert-reproduction-scores}). Expert ratings were strongly associated with the continuous artifact scores (Spearman $\rho = 0.76$, $p = 1.3 \times 10^{-5}$; Supplementary Figure~\ref{fig:supp-expert-score}). Expert feedback was additionally used to refine the task prompts, method guidance, and grading for these tasks.

\begin{table}[!ht]
\centering
\caption{\textbf{Expert reproduction rating definitions.} Domain experts assigned each reviewed artifact an integer rating from 1 to 5 using these criteria.}
\label{tab:expert-reproduction-scores}
\begin{tabular}{@{}c p{0.82\textwidth}@{}}
\toprule
Rating & Definition \\
\midrule
1 & \textbf{Failed reproduction.} The artifact is missing, unreadable, empty, synthetic, unrelated to the requested output, or so different from the published artifact that it provides no meaningful reproduction. \\
2 & \textbf{Weak reproduction.} The artifact is related to the requested analysis, but most of the published result was not recovered. Only a small subset, coarse pattern, or structural resemblance remains, and the main scientific interpretation is unreliable. \\
3 & \textbf{Partial reproduction.} The agent recovered a recognizable portion of the result, but material differences affect important rows, columns, values, labels, or conclusions. Some major components are correct, and others are missing or wrong. The same biological conclusions could not be drawn from the agent's artifact as from the published artifact. \\
4 & \textbf{Mostly accurate.} The main result was reproduced, but there are limited discrepancies in coverage, values, labels, thresholds, or implementation. The differences are real and should be documented, but they do not overturn the artifact's main interpretation. \\
5 & \textbf{Faithful reproduction.} The agent reproduced the artifact's scientific content. Column and row coverage is complete or nearly complete, shared numerical values or labels agree closely, and the same conclusions follow from both files. Any remaining differences are cosmetic or too small to affect interpretation. \\
\bottomrule
\end{tabular}
\end{table}

\subsection{Process-judge failure modes}

\begin{longtable}{@{}p{0.27\textwidth}p{0.68\textwidth}@{}}
\caption{Definitions of the closed-vocabulary failure modes assigned by the process judge.}\label{tab:failure-modes}\\
\toprule
Failure mode & Description \\
\midrule
\endfirsthead
\toprule
Failure mode & Description \\
\midrule
\endhead
Environment setup failure & The agent could not get required tools, packages, references, or paths into a usable state. The tag is used when that unresolved setup problem prevents or materially blocks the intended analysis. \\
Input misinterpretation & The agent misread the staged inputs, confused sample identities, or applied an analysis intended for a different data type. It captures an incorrect interpretation of the data supplied to the run, rather than a later choice of analysis parameters. \\
Wrong method substitution & The agent used a method materially different from the one required by Method Guidance, such as a different tool family or statistical approach. The substitution was not permitted by the prompt. \\
Method misconfigured & The agent used the required method but chose parameters, contrasts, references, or filters that materially diverged from Method Guidance. The method itself is correct, which distinguishes this failure from wrong method substitution. \\
Incomplete data & The agent analyzed fewer samples, conditions, or features than the prompt specified. The omitted data leave the analysis incomplete even if the reduced analysis ran successfully. \\
Output format violation & The agent produced an artifact that did not match the Required Outputs schema. Violations include the wrong filename, location, index column, required columns, data types, or units. \\
Synthetic or placeholder output & The agent wrote synthetic, dummy, empty, or formulaic values instead of deriving output from the real staged data. The tag applies whether fabrication is explicit or used to conceal a failed or incomplete analysis. \\
Premature termination & The agent stopped before writing all required outputs because it exhausted a turn budget or time limit, or because it declared completion before the work was finished. At least one required artifact therefore remained unwritten. \\
Inefficient analysis & The agent failed to use available time productively. Examples include ordering work inefficiently, missing obvious parallelization, omitting available parallel flags, or unnecessarily repeating completed steps. \\
Repetitive retry loop & The agent repeatedly retried substantially the same failing command or approach. The retries lacked a meaningful diagnosis or change in strategy. \\
\bottomrule
\end{longtable}

\subsection{Example task prompt}
\label{app:example-prompt}

The following is the complete prompt for the task shown in Figure~\ref{fig:example}:

\begin{lstlisting}[
  basicstyle=\fontfamily{cmtt}\selectfont\scriptsize,
  breaklines=true,
  breakatwhitespace=false,
  columns=fullflexible,
  keepspaces=true,
  showstringspaces=false,
  literate={—}{{\textemdash}}1
]
# Analysis Task

You are a computational biologist performing an analysis end to end. Complete the entire task and produce the requested output files.

## Working Directory

- Input data and reference files are provided under `./data`. You may use other data from the web as needed.
- The `./data` directory is read-only. If you need to modify, decompress, index, or otherwise rewrite an input file, copy it to `./work` first.
- Write all analysis outputs under `./outputs`.
- You may create intermediate files under `./work` if useful.

## Runtime Limit

You have a maximum wall-clock budget of `86400` seconds for this task. The run will be terminated when this budget is reached.

## Already Available Software

The sandbox image already includes the following general-purpose software:

- Build tooling: `build-essential`, `gfortran`, `make`, `cmake`, `pkg-config`. Compilers and build helpers for source package installation.
- Archive and download utilities: `wget`, `curl`, `aria2`, `rsync`, `unzip`, `zip`, `tar`, `pigz`, `jq`. Common command-line tools for fetching, unpacking, and inspecting files.
- Core development libraries: `libcurl4-openssl-dev`, `libssl-dev`, `libxml2-dev`, `zlib1g-dev`, `libbz2-dev`, `liblzma-dev`, `libzstd-dev`, `libhdf5-dev`, `libgit2-dev`. Header libraries used by common Python, R, and bioinformatics packages.
- Plot, font, and image libraries: `libpng-dev`, `libjpeg-dev`, `libtiff-dev`, `libfreetype6-dev`, `libfontconfig1-dev`, `libharfbuzz-dev`, `libfribidi-dev`, `libcairo2-dev`. Graphics libraries used by plotting and reporting packages.
- Java: `openjdk-17-jre-headless`. Java runtime for Java-based bioinformatics tools and workflow helpers.
- Python: `python3`, `python3-pip`, `uv`, `numpy`, `pandas`, `scipy`, `scikit-learn`, `statsmodels`, `pyarrow`, `h5py`, `matplotlib`, `biopython`, `pysam`. Python runtime, package installer, scientific stack, and core bioinformatics libraries.
- R: `r-base`, `r-base-dev`, `BiocManager`, `data.table`, `dplyr`, `readr`, `tidyr`, `ggplot2`, `Matrix`, `remotes`, `pak`. R runtime, build support, package installers, and common data-wrangling packages.
- Conda-compatible package management: `micromamba`, `mamba`. Micromamba is configured with conda-forge, bioconda, and defaults channels.
- 10x Genomics single-cell tools: `cellranger 6.0.2`, `cellranger-arc 2.0.2`.

Method-specific analysis packages are not preinstalled unless listed here; install the tools and versions required by the method guidance when needed.



## Network and External Resources

You have network access through an allow-list proxy. Use it freely to install packages and pull canonical reference data and method tools. Lean toward fetching what you need rather than guessing.

### Pre-approved (just install or fetch, no approval call needed)

- **Python packages**: `pip install <pkg>` from PyPI (`pypi.org`, `files.pythonhosted.org`).
- **Conda packages**: `conda install -c conda-forge -c bioconda <pkg>` or `mamba install ...` (`conda.anaconda.org`, `repo.anaconda.com`, `anaconda.org`).
- **R packages**: `install.packages("<pkg>")` from CRAN (`cran.r-project.org`, `cloud.r-project.org`, `packagemanager.posit.co`) and `BiocManager::install("<pkg>")` from Bioconductor (`bioconductor.org`).
- **OS packages**: `apt-get install <pkg>` from Debian and Ubuntu mirrors.
- **Other language registries**: npm (`registry.npmjs.org`), Cargo (`crates.io`), Go modules (`proxy.golang.org`), RubyGems.
- **Astral tools**: `uv`, `ruff`, etc. (`astral.sh`).

Always try the package manager first. Only escalate to `request_web_access` if the tool or data you need is not available through a packaging system.

### Approved on request via `request_web_access(url, reason)`

When you need a resource outside the pre-approved infrastructure, call `request_web_access` with the exact URL and a short justification. The following classes of URLs are routinely approved:

- **Method-tool source code on GitHub/GitLab/Bitbucket**: a repository root like `https://github.com/<org>/<repo>`, a release tarball, or a raw file from a tool you need to run. Bare repository URLs are cloned into your workspace; specific release URLs are fetched as files.
- **Canonical public reference data** from primary providers, for example:
    - Ensembl FTP (`ftp.ensembl.org`) for genome FASTA, GTF/GFF, cDNA.
    - GENCODE (`ftp.ebi.ac.uk/pub/databases/gencode`).
    - UCSC goldenPath (`hgdownload.soe.ucsc.edu`) for FASTA, chrom sizes, blacklists.
    - NCBI Genome / RefSeq FTP (`ftp.ncbi.nlm.nih.gov`).
    - UniProt (`rest.uniprot.org`, `ftp.uniprot.org`) for protein FASTA and metadata.
    - JASPAR, MSigDB, Pfam, InterPro, KEGG REST, OBO Foundry, and similar canonical bioinformatics resources.
- **Tool documentation pages, vignettes, manuals, READMEs, and API references** when you need to look up correct usage.
- **Specific raw-read accession files** from SRA/ENA when the staged data does not already include them (only the specific file URL — not a search page).

When the broker approves, branch on the response:

- `delivery: "allowlist"` — the host was added to the proxy allowlist; rerun your original fetch command.
- `delivery: "workspace_file"` — the broker downloaded the resource for you; read it from `local_path` (lands under `/workspace/approved_web/`).

### Never approved — do not spend budget asking

- PDFs, abstracts, supplements, source-data workbooks, extended-data files, or any other answer-bearing artifact for published papers.
- Processed result matrices, normalized matrices, cell-type labels, cluster assignments, differential-expression tables, peak lists, or similar precomputed outputs for published datasets — even if hosted on GEO, ArrayExpress, Synapse, CellxGene, Zenodo, Figshare, OSF, Dryad, Google Drive, or Dropbox.
- Broad search-engine queries (Google, Bing) containing paper-specific terms or accessions.

## Conduct

- Use the tools, parameters, thresholds, and databases named under "Method Guidance".
- When a detail is unspecified, choose a reasonable default that is consistent with the named tools and document it briefly in a top-level comment of the analysis script that uses it.
- Prefer scripted, reproducible steps over interactive ones. Pin software versions when you can.
- Contain all your work to `./data`, `./work`, and `./outputs`.
- Be efficient with your use of tools and resources. Do not poll long running commands unnecessarily frequently. You can run long running commands in the background or sleep between checks.

## Output Guidance

The "Required Outputs" section below specifies the exact files to produce. For each file:

- Write to the exact relative path given.
- Use the exact format requested (CSV, TSV, BED, etc.).
- Include a header row when the format is tabular.
- Use the exact column names listed when names are given. When only a description is given, choose stable, descriptive column names that match the description.
- Place the row-key column as the first column.
- Include one row per requested entity. Do not add summary, totals, or commentary rows.
- Do not write notes, narrative, or markdown into data files.
- Use UTF-8 text encoding and Unix newlines.

### `METHODS.md` (required)

Write `./outputs/METHODS.md` describing how each required artifact was produced, the tools and parameters used, and the data sources and intermediate files created. Write this in the style of the methods section of a scientific paper. Use one `## {artifact_id}` section per required artifact. For each artifact, include the tools, versions, parameters, input files under `./data`, intermediate files under `./work`, and output file written under `./outputs`. Explicitly disclose any substitutions, skipped steps, different tool versions, fallback methods, failed commands, partial outputs, or assumptions. This file should be sufficiently detailed for another researcher to exactly reproduce the analysis from scratch.

## Research Objective

Characterize how loss of TCER-1 changes endogenous small-RNA regulation and mRNA expression in *Caenorhabditis elegans*, with emphasis on WAGO-associated 22G-RNAs and their target transcripts during adult growth on non-pathogenic *E. coli* OP50.

`./data` contains raw sequencing inputs for three biological replicates each of wild-type N2, `tcer-1(glm27)`, and `tcer-1(tm1452)` gravid adult animals. The mRNA libraries are rRNA-depleted, directional, paired-end 150-cycle RNA-seq FASTQs, and the small-RNA libraries are polyphosphatase-treated, 16-30 nt size-selected, single-end 75-cycle FASTQs. `./data/metadata` contains a sample sheet with condition, replicate, library type, and layout assignments, short pipeline notes, an exact WormBase WS279 PRJNA13758 mRNA/RSEM reference bundle under `./data/metadata/reference/wormbase_ws279_mrna/`, and a Montgomery Lab tinyRNA WS279 reference bundle under `./data/metadata/reference/montgomery_tinyrna_ws279/`. Quantify mRNA expression and small-RNA abundance, compare each `tcer-1` allele with N2, identify concordant at-least-twofold changes across both alleles, and integrate WAGO-class 22G-RNA changes with paired mRNA changes for each allele.

## Method Guidance

Use the staged sample sheet to assign libraries to `N2`, `tcer-1(glm27)`, and `tcer-1(tm1452)` groups, keeping three biological replicates per group and separate mRNA-seq and small-RNA-seq designs. Treat N2 as the reference condition for all differential analyses. For every log2 fold change, use mutant over N2 direction: positive values indicate higher expression or abundance in the mutant, and negative values indicate higher expression or abundance in N2.

For mRNA-seq, trim each paired library with `fastp` using `fastp -w 16 -q 30 -u 70 -l 30 -r -W 4 -M 20`. Build the `RSEM v1.3.1` reference from the exact staged WormBase WS279 PRJNA13758 mRNA files: `./data/metadata/reference/wormbase_ws279_mrna/c_elegans.PRJNA13758.WS279.genomic.fa.gz` and `./data/metadata/reference/wormbase_ws279_mrna/c_elegans.PRJNA13758.WS279.canonical_geneset.gtf.gz`, using `rsem-prepare-reference --gtf` with `--star` so WormBase gene IDs are the gene-level identifiers, and quantify with `STAR v2.7.10b` as the aligner. Summarize to the gene-level rows emitted by RSEM expected-count quantification, retaining WormBase gene identifiers and gene names where available. Analyze the count matrix with `DESeq2 v1.50.2`, using median-of-ratios size-factor normalization, a condition-only design with N2 as the reference level, Wald-statistic inference, and Benjamini-Hochberg adjusted p-values. Extract the two mutant-versus-N2 contrasts, report DESeq2-normalized counts for the RSEM gene-level row universe, and use absolute log2 fold change of at least 1 with adjusted p-value below 0.05 when defining genes changed concordantly across both mutant alleles.

For small-RNA-seq, process the nine single-end libraries with the `tinyRNA` pipeline using default settings against the staged Montgomery Lab WS279 tinyRNA bundle: `./data/metadata/reference/montgomery_tinyrna_ws279/c_elegans.PRJNA13758.WS279.genomic.fa.gz`, `./data/metadata/reference/montgomery_tinyrna_ws279/sRNAs_WS279.gff3`, and `./data/metadata/reference/montgomery_tinyrna_ws279/features_cel_v1.5.csv`. Use the library design as 16-30 nt, polyphosphatase-treated small RNA, and define 22G-RNAs as antisense-strand 21-23 nt reads containing a 5-prime G. Carry the tinyRNA feature universe through downstream tables, including miRNA MIMAT rows, sequence-level 23H-RNA rows, and class-specific WBGene 22G rows. Run DESeq2 within `tinyRNA` with the same N2 reference, mutant-versus-N2 contrast direction, Wald-statistic inference, and Benjamini-Hochberg adjustment. Use DESeq2 size-factor normalized small-RNA counts and the same twofold and adjusted-p-value criteria for concordant allele-level changes. When a small-RNA feature maps to the same gene in multiple classes, keep the feature class and feature name attached so distinct small-RNA classes remain separate.

For the integrated 22G-RNA/mRNA analysis, run `RNA-integrate` on the count tables generated from the `tinyRNA` and `RSEM` workflows, with metadata marking N2 as the control condition and both `tcer-1(tm1452)` versus N2 and `tcer-1(glm27)` versus N2 as cross-comparisons. Use `./data/metadata/reference/montgomery_tinyrna_ws279/sRNAs_WS279.gff3` as the WAGO target-pairing source and `./data/metadata/reference/montgomery_tinyrna_ws279/features_cel_v1.5.csv` as the tinyRNA class-rule source for the gene class/name information required by `RNA-integrate`. Restrict integrated summaries to annotated WAGO 22G-RNA target genes. For each allele separately, compare the small-RNA and mRNA log2 mutant-over-N2 effects, and derive anticorrelated groups with small RNA down and mRNA up more than twofold, as well as the reciprocal small RNA up and mRNA down more than twofold, using adjusted p-value below 0.05 for significance classification.

## Required Outputs

### `mrna_normalized_counts`

- Path: `outputs/mrna_normalized_counts.csv`
- Format: `csv`
- Description: Geometric mean/DESeq2-style normalized mRNA-seq counts for C. elegans gene features across wild-type N2, tcer-1(glm27), and tcer-1(tm1452) biological replicates.
- Row key: `Feature_ID`
- Rows: One row per gene-level RSEM v1.3.1 expected-count entry identified by WormBase gene ID after quantifying against the exact staged WormBase WS279 PRJNA13758 RSEM reference built from `wormbase_ws279_mrna/c_elegans.PRJNA13758.WS279.genomic.fa.gz` and `wormbase_ws279_mrna/c_elegans.PRJNA13758.WS279.canonical_geneset.gtf.gz`.
- Example row identifiers: `WBGene00000001`, `WBGene00000002`, `WBGene00000003`
- Columns: Identifier and gene-name columns followed by normalized count columns for three N2, three tcer-1(glm27), and three tcer-1(tm1452) replicates.
- Required columns: `Feature_ID`, `Feature_Name`, `N2_rep_1`, `N2_rep_2`, `N2_rep_3`, `tcer-1(glm27)_rep_1`, `tcer-1(glm27)_rep_2`, `tcer-1(glm27)_rep_3`, `tcer-1(tm1452)_rep_1`, `tcer-1(tm1452)_rep_2`, `tcer-1(tm1452)_rep_3`

### `srna_normalized_counts`

- Path: `outputs/srna_normalized_counts.csv`
- Format: `csv`
- Description: Normalized small-RNA abundance matrix for C. elegans small-RNA features across wild-type N2, tcer-1(glm27), and tcer-1(tm1452) biological replicates. Use Feature_Key as Feature_ID|Classifier|Feature_Name because Feature_ID alone is not unique for this small-RNA table.
- Row key: `Feature_Key`
- Rows: One row per tinyRNA-tested small-RNA feature emitted from the staged Montgomery Lab WS279 tinyRNA bundle, keyed by the exact concatenation Feature_ID|Classifier|Feature_Name. The feature universe includes miRNA MIMAT rows, sequence-level 23H-RNA rows, and class-specific WBGene 22G rows.
- Example row identifiers: `AAAAAAGAACTGAAGAGAGTG|23H-RNA|C05E7.1, C05E7.t3`, `MIMAT0000001|miRNA|let-7-5p`, `WBGene00001335|WAGO Class 22G|F35A5.8`
- Columns: Composite key and feature annotation columns followed by normalized count columns for three N2, three tcer-1(glm27), and three tcer-1(tm1452) replicates.
- Required columns: `Feature_Key`, `Feature_ID`, `Classifier`, `Feature_Name`, `N2_rep_1`, `N2_rep_2`, `N2_rep_3`, `tcer-1(glm27)_rep_1`, `tcer-1(glm27)_rep_2`, `tcer-1(glm27)_rep_3`, `tcer-1(tm1452)_rep_1`, `tcer-1(tm1452)_rep_2`, `tcer-1(tm1452)_rep_3`

### `srna_differential_expression_all`

- Path: `outputs/srna_differential_expression_all.csv`
- Format: `csv`
- Description: All tested small-RNA features from the DESeq2-style analysis. Small-RNA differential expression values compare each tcer-1 mutant against wild-type N2; log2FC_tm1452_vs_wt is log2(tcer-1(tm1452) / wild type) and log2FC_glm27_vs_wt is log2(tcer-1(glm27) / wild type), so positive values indicate higher abundance in the mutant and negative values indicate higher abundance in wild type.
- Row key: `Feature_Key`
- Rows: One row per tinyRNA-tested small-RNA feature emitted from the staged Montgomery Lab WS279 tinyRNA bundle, keyed by the exact concatenation Feature_ID|Classifier|Feature_Name. The feature universe includes miRNA MIMAT rows, sequence-level 23H-RNA rows, and class-specific WBGene 22G rows.
- Example row identifiers: `AAAAAAGAACTGAAGAGAGTG|23H-RNA|C05E7.1, C05E7.t3`, `MIMAT0000001|miRNA|let-7-5p`, `WBGene00001335|WAGO Class 22G|F35A5.8`
- Columns: Composite row key, original feature annotations, mean reads, adjusted p values, and allele-specific log2 fold changes with mutant-over-wild-type direction.
- Required columns: `Feature_Key`, `Feature_ID`, `Classifier`, `Feature_Name`, `mean_reads`, `log2FC_tm1452_vs_wt`, `padj_tm1452_vs_wt`, `log2FC_glm27_vs_wt`, `padj_glm27_vs_wt`

### `srna_down_twofold_both_mutants`

- Path: `outputs/srna_down_twofold_both_mutants.csv`
- Format: `csv`
- Description: Small-RNA features reported as down at least two-fold in both tcer-1 mutants. Small-RNA differential expression values compare each tcer-1 mutant against wild-type N2; log2FC_tm1452_vs_wt is log2(tcer-1(tm1452) / wild type) and log2FC_glm27_vs_wt is log2(tcer-1(glm27) / wild type), so positive values indicate higher abundance in the mutant and negative values indicate higher abundance in wild type.
- Row key: `Feature_Key`
- Rows: One row per tinyRNA-tested small-RNA feature from the staged Montgomery Lab WS279 tinyRNA bundle that meets this artifact's two-allele fold-change and adjusted-p-value rule, keyed by Feature_ID|Classifier|Feature_Name.
- Example row identifiers: `AAAAAAGAACTGAAGAGAGTG|23H-RNA|C05E7.1, C05E7.t3`, `MIMAT0000001|miRNA|let-7-5p`, `WBGene00001335|WAGO Class 22G|F35A5.8`
- Columns: Composite row key, original feature annotations, mean reads, adjusted p values, and allele-specific log2 fold changes with mutant-over-wild-type direction.
- Required columns: `Feature_Key`, `Feature_ID`, `Classifier`, `Feature_Name`, `mean_reads`, `log2FC_tm1452_vs_wt`, `padj_tm1452_vs_wt`, `log2FC_glm27_vs_wt`, `padj_glm27_vs_wt`

### `srna_up_twofold_both_mutants`

- Path: `outputs/srna_up_twofold_both_mutants.csv`
- Format: `csv`
- Description: Small-RNA features reported as up at least two-fold in both tcer-1 mutants. Small-RNA differential expression values compare each tcer-1 mutant against wild-type N2; log2FC_tm1452_vs_wt is log2(tcer-1(tm1452) / wild type) and log2FC_glm27_vs_wt is log2(tcer-1(glm27) / wild type), so positive values indicate higher abundance in the mutant and negative values indicate higher abundance in wild type.
- Row key: `Feature_Key`
- Rows: One row per tinyRNA-tested small-RNA feature from the staged Montgomery Lab WS279 tinyRNA bundle that meets this artifact's two-allele fold-change and adjusted-p-value rule, keyed by Feature_ID|Classifier|Feature_Name.
- Example row identifiers: `AAAAAAGAACTGAAGAGAGTG|23H-RNA|C05E7.1, C05E7.t3`, `MIMAT0000001|miRNA|let-7-5p`, `WBGene00001335|WAGO Class 22G|F35A5.8`
- Columns: Composite row key, original feature annotations, mean reads, adjusted p values, and allele-specific log2 fold changes with mutant-over-wild-type direction.
- Required columns: `Feature_Key`, `Feature_ID`, `Classifier`, `Feature_Name`, `mean_reads`, `log2FC_tm1452_vs_wt`, `padj_tm1452_vs_wt`, `log2FC_glm27_vs_wt`, `padj_glm27_vs_wt`

### `mrna_differential_expression_all`

- Path: `outputs/mrna_differential_expression_all.csv`
- Format: `csv`
- Description: All tested mRNA gene features from the DESeq2-style analysis. mRNA differential expression values compare each tcer-1 mutant against wild-type N2; log2FC_tm1452_vs_wt is log2(tcer-1(tm1452) / wild type) and log2FC_glm27_vs_wt is log2(tcer-1(glm27) / wild type), so positive values indicate higher expression in the mutant and negative values indicate higher expression in wild type.
- Row key: `Feature_ID`
- Rows: One row per gene-level RSEM/DESeq2-tested mRNA feature identified by WormBase gene ID after quantifying against the exact staged WormBase WS279 PRJNA13758 RSEM reference.
- Example row identifiers: `WBGene00000001`, `WBGene00000002`, `WBGene00000003`
- Columns: Gene identifiers, gene names, WAGO/CSR-1 target annotation, normalized replicate counts, adjusted p values, and mutant-over-wild-type log2 fold changes for each tcer-1 allele.
- Required columns: `Feature_ID`, `Feature_Name`, `WAGO_CSR1_target`, `WT_rep_1`, `WT_rep_2`, `WT_rep_3`, `tm1452_rep_1`, `tm1452_rep_2`, `tm1452_rep_3`, `log2FC_tm1452_vs_wt`, `padj_tm1452_vs_wt`, `glm27_rep_1`, `glm27_rep_2`, `glm27_rep_3`, `log2FC_glm27_vs_wt`, `padj_glm27_vs_wt`

### `mrna_up_twofold_both_mutants`

- Path: `outputs/mrna_up_twofold_both_mutants.csv`
- Format: `csv`
- Description: mRNA features reported as up at least two-fold in both tcer-1 mutants. mRNA differential expression values compare each tcer-1 mutant against wild-type N2; log2FC_tm1452_vs_wt is log2(tcer-1(tm1452) / wild type) and log2FC_glm27_vs_wt is log2(tcer-1(glm27) / wild type), so positive values indicate higher expression in the mutant and negative values indicate higher expression in wild type.
- Row key: `Feature_ID`
- Rows: One row per gene-level RSEM/DESeq2-tested WormBase gene ID with log2 fold change at least 1 and adjusted p-value below 0.05 in both tcer-1 mutant contrasts.
- Example row identifiers: `WBGene00000001`, `WBGene00000002`, `WBGene00000003`
- Columns: Gene identifiers, gene names, WAGO/CSR-1 target annotation, normalized replicate counts, adjusted p values, and mutant-over-wild-type log2 fold changes for each tcer-1 allele.
- Required columns: `Feature_ID`, `Feature_Name`, `WAGO_CSR1_target`, `WT_rep_1`, `WT_rep_2`, `WT_rep_3`, `tm1452_rep_1`, `tm1452_rep_2`, `tm1452_rep_3`, `log2FC_tm1452_vs_wt`, `padj_tm1452_vs_wt`, `glm27_rep_1`, `glm27_rep_2`, `glm27_rep_3`, `log2FC_glm27_vs_wt`, `padj_glm27_vs_wt`

### `mrna_down_twofold_both_mutants`

- Path: `outputs/mrna_down_twofold_both_mutants.csv`
- Format: `csv`
- Description: mRNA features reported as down at least two-fold in both tcer-1 mutants. mRNA differential expression values compare each tcer-1 mutant against wild-type N2; log2FC_tm1452_vs_wt is log2(tcer-1(tm1452) / wild type) and log2FC_glm27_vs_wt is log2(tcer-1(glm27) / wild type), so positive values indicate higher expression in the mutant and negative values indicate higher expression in wild type.
- Row key: `Feature_ID`
- Rows: One row per gene-level RSEM/DESeq2-tested WormBase gene ID with log2 fold change at most -1 and adjusted p-value below 0.05 in both tcer-1 mutant contrasts.
- Example row identifiers: `WBGene00000001`, `WBGene00000002`, `WBGene00000003`
- Columns: Gene identifiers, gene names, WAGO/CSR-1 target annotation, normalized replicate counts, adjusted p values, and mutant-over-wild-type log2 fold changes for each tcer-1 allele.
- Required columns: `Feature_ID`, `Feature_Name`, `WAGO_CSR1_target`, `WT_rep_1`, `WT_rep_2`, `WT_rep_3`, `tm1452_rep_1`, `tm1452_rep_2`, `tm1452_rep_3`, `log2FC_tm1452_vs_wt`, `padj_tm1452_vs_wt`, `glm27_rep_1`, `glm27_rep_2`, `glm27_rep_3`, `log2FC_glm27_vs_wt`, `padj_glm27_vs_wt`

### `integrative_tm1452_all`

- Path: `outputs/integrative_tm1452_all.csv`
- Format: `csv`
- Description: All WAGO-target features from the tcer-1(tm1452) integrated 22G-RNA/mRNA analysis. Integrated WAGO-target 22G-RNA and mRNA values compare tcer-1(tm1452) against wild-type N2; the sRNA and mRNA log2FC columns are log2(tcer-1(tm1452) / wild type), so positive values indicate higher abundance or expression in the mutant and negative values indicate higher abundance or expression in wild type.
- Row key: `Feature_ID`
- Rows: One row per WAGO-target C. elegans gene feature identified by WormBase gene ID, using RNA-integrate with WAGO target pairings derived from the staged WS279 sRNAs_WS279.gff3 and features_cel_v1.5.csv files.
- Example row identifiers: `WBGene00000022`, `WBGene00000034`, `WBGene00000075`
- Columns: Gene identifiers, gene names, mean sRNA and mRNA values, adjusted p values, and mutant-over-wild-type sRNA and mRNA log2 fold changes for tcer-1(tm1452).
- Required columns: `Feature_ID`, `Feature_Name`, `srna_mean_tm1452`, `srna_log2FC_tm1452_vs_wt`, `srna_padj_tm1452_vs_wt`, `mrna_mean_tm1452`, `mrna_log2FC_tm1452_vs_wt`, `mrna_padj_tm1452_vs_wt`

### `integrative_tm1452_srna_down_mrna_up`

- Path: `outputs/integrative_tm1452_srna_down_mrna_up.csv`
- Format: `csv`
- Description: WAGO-target features reported with small RNAs down more than two-fold and mRNA up more than two-fold in tcer-1(tm1452). Integrated WAGO-target 22G-RNA and mRNA values compare tcer-1(tm1452) against wild-type N2; the sRNA and mRNA log2FC columns are log2(tcer-1(tm1452) / wild type), so positive values indicate higher abundance or expression in the mutant and negative values indicate higher abundance or expression in wild type.
- Row key: `Feature_ID`
- Rows: One row per WAGO-target WormBase gene ID from the RNA-integrate result with small RNA log2 fold change at most -1, mRNA log2 fold change at least 1, and adjusted p-value below 0.05 for tcer-1(tm1452).
- Example row identifiers: `WBGene00000022`, `WBGene00000034`, `WBGene00000075`
- Columns: Gene identifiers, gene names, mean sRNA and mRNA values, adjusted p values, and mutant-over-wild-type sRNA and mRNA log2 fold changes for tcer-1(tm1452).
- Required columns: `Feature_ID`, `Feature_Name`, `srna_mean_tm1452`, `srna_log2FC_tm1452_vs_wt`, `srna_padj_tm1452_vs_wt`, `mrna_mean_tm1452`, `mrna_log2FC_tm1452_vs_wt`, `mrna_padj_tm1452_vs_wt`

### `integrative_tm1452_srna_up_mrna_down`

- Path: `outputs/integrative_tm1452_srna_up_mrna_down.csv`
- Format: `csv`
- Description: WAGO-target features reported with small RNAs up more than two-fold and mRNA down more than two-fold in tcer-1(tm1452). Integrated WAGO-target 22G-RNA and mRNA values compare tcer-1(tm1452) against wild-type N2; the sRNA and mRNA log2FC columns are log2(tcer-1(tm1452) / wild type), so positive values indicate higher abundance or expression in the mutant and negative values indicate higher abundance or expression in wild type.
- Row key: `Feature_ID`
- Rows: One row per WAGO-target WormBase gene ID from the RNA-integrate result with small RNA log2 fold change at least 1, mRNA log2 fold change at most -1, and adjusted p-value below 0.05 for tcer-1(tm1452).
- Example row identifiers: `WBGene00000022`, `WBGene00000034`, `WBGene00000075`
- Columns: Gene identifiers, gene names, mean sRNA and mRNA values, adjusted p values, and mutant-over-wild-type sRNA and mRNA log2 fold changes for tcer-1(tm1452).
- Required columns: `Feature_ID`, `Feature_Name`, `srna_mean_tm1452`, `srna_log2FC_tm1452_vs_wt`, `srna_padj_tm1452_vs_wt`, `mrna_mean_tm1452`, `mrna_log2FC_tm1452_vs_wt`, `mrna_padj_tm1452_vs_wt`

### `integrative_glm27_all`

- Path: `outputs/integrative_glm27_all.csv`
- Format: `csv`
- Description: All WAGO-target features from the tcer-1(glm27) integrated 22G-RNA/mRNA analysis. Integrated WAGO-target 22G-RNA and mRNA values compare tcer-1(glm27) against wild-type N2; the sRNA and mRNA log2FC columns are log2(tcer-1(glm27) / wild type), so positive values indicate higher abundance or expression in the mutant and negative values indicate higher abundance or expression in wild type.
- Row key: `Feature_ID`
- Rows: One row per WAGO-target C. elegans gene feature identified by WormBase gene ID, using RNA-integrate with WAGO target pairings derived from the staged WS279 sRNAs_WS279.gff3 and features_cel_v1.5.csv files.
- Example row identifiers: `WBGene00000022`, `WBGene00000034`, `WBGene00000075`
- Columns: Gene identifiers, gene names, mean sRNA and mRNA values, adjusted p values, and mutant-over-wild-type sRNA and mRNA log2 fold changes for tcer-1(glm27).
- Required columns: `Feature_ID`, `Feature_Name`, `srna_mean_glm27`, `srna_log2FC_glm27_vs_wt`, `srna_padj_glm27_vs_wt`, `mrna_mean_glm27`, `mrna_log2FC_glm27_vs_wt`, `mrna_padj_glm27_vs_wt`

### `integrative_glm27_srna_down_mrna_up`

- Path: `outputs/integrative_glm27_srna_down_mrna_up.csv`
- Format: `csv`
- Description: WAGO-target features reported with small RNAs down more than two-fold and mRNA up more than two-fold in tcer-1(glm27). Integrated WAGO-target 22G-RNA and mRNA values compare tcer-1(glm27) against wild-type N2; the sRNA and mRNA log2FC columns are log2(tcer-1(glm27) / wild type), so positive values indicate higher abundance or expression in the mutant and negative values indicate higher abundance or expression in wild type.
- Row key: `Feature_ID`
- Rows: One row per WAGO-target WormBase gene ID from the RNA-integrate result with small RNA log2 fold change at most -1, mRNA log2 fold change at least 1, and adjusted p-value below 0.05 for tcer-1(glm27).
- Example row identifiers: `WBGene00000022`, `WBGene00000034`, `WBGene00000075`
- Columns: Gene identifiers, gene names, mean sRNA and mRNA values, adjusted p values, and mutant-over-wild-type sRNA and mRNA log2 fold changes for tcer-1(glm27).
- Required columns: `Feature_ID`, `Feature_Name`, `srna_mean_glm27`, `srna_log2FC_glm27_vs_wt`, `srna_padj_glm27_vs_wt`, `mrna_mean_glm27`, `mrna_log2FC_glm27_vs_wt`, `mrna_padj_glm27_vs_wt`

### `integrative_glm27_srna_up_mrna_down`

- Path: `outputs/integrative_glm27_srna_up_mrna_down.csv`
- Format: `csv`
- Description: WAGO-target features reported with small RNAs up more than two-fold and mRNA down more than two-fold in tcer-1(glm27). Integrated WAGO-target 22G-RNA and mRNA values compare tcer-1(glm27) against wild-type N2; the sRNA and mRNA log2FC columns are log2(tcer-1(glm27) / wild type), so positive values indicate higher abundance or expression in the mutant and negative values indicate higher abundance or expression in wild type.
- Row key: `Feature_ID`
- Rows: One row per WAGO-target WormBase gene ID from the RNA-integrate result with small RNA log2 fold change at least 1, mRNA log2 fold change at most -1, and adjusted p-value below 0.05 for tcer-1(glm27).
- Example row identifiers: `WBGene00000022`, `WBGene00000034`, `WBGene00000075`
- Columns: Gene identifiers, gene names, mean sRNA and mRNA values, adjusted p values, and mutant-over-wild-type sRNA and mRNA log2 fold changes for tcer-1(glm27).
- Required columns: `Feature_ID`, `Feature_Name`, `srna_mean_glm27`, `srna_log2FC_glm27_vs_wt`, `srna_padj_glm27_vs_wt`, `mrna_mean_glm27`, `mrna_log2FC_glm27_vs_wt`, `mrna_padj_glm27_vs_wt`
\end{lstlisting}

\end{document}